\documentclass[letterpaper, 10 pt, journal, twoside]{IEEEtran}
\usepackage{amsmath,amsfonts}
\usepackage{algorithmic}
\usepackage{array}
\usepackage{textcomp}
\usepackage{stfloats}
\usepackage{url}
\usepackage{verbatim}
\usepackage{graphicx}
\def\BibTeX{{\rm B\kern-.05em{\sc i\kern-.025em b}\kern-.08em
    T\kern-.1667em\lower.7ex\hbox{E}\kern-.125emX}}
\usepackage{balance}

\usepackage{graphicx}

\usepackage{gensymb}
\usepackage{multirow}
\usepackage{multicol}
\usepackage{tabularx}
\usepackage{dsfont}
\usepackage{url}
\usepackage{subcaption}
\usepackage{bm}
\usepackage{wrapfig}
\usepackage{caption}
\usepackage{mathrsfs}
\usepackage{float}
\usepackage{booktabs}
\usepackage{dcolumn}
\usepackage{pifont}

\usepackage{amssymb}
\usepackage{hyperref}
\usepackage{amsmath}
\usepackage{natbib} 
\usepackage[ruled,linesnumbered]{algorithm2e}

\usepackage{xcolor}

\usepackage{colortbl}
\usepackage[most]{tcolorbox}
\usepackage{listings}
\usepackage{upquote}
\tcbuselibrary{listings, breakable}
\newtcolorbox{response}[1][]{
  colback=gray!5,
  colframe=black,
  fonttitle=\bfseries,
  coltitle=black,
  }
\newtcblisting{responselong}[1][]{
  enhanced,
  breakable,                
  colback=gray!5,
  colframe=black,
  fonttitle=\bfseries,
  coltitle=white,
  listing only,             
  listing options={         
    basicstyle=\small\ttfamily,
    breaklines=true,        
  },
  #1                        
}

\newtcolorbox{responselongimg}[1][]{
  enhanced,
  breakable,
  colback=gray!5,
  colframe=black,
  fonttitle=\bfseries,
  coltitle=white,
  #1
}

\usepackage[utf8]{inputenc} 
\usepackage[T1]{fontenc}    
\usepackage{hyperref}       
\usepackage{url}            
\usepackage{booktabs}       
\usepackage{amsfonts}       
\usepackage{nicefrac}       
\usepackage{microtype}      
\usepackage{xcolor}         
\usepackage{multirow}

\usepackage{graphicx}

\usepackage{multicol}
\usepackage{colortbl}
\definecolor{mygray}{gray}{.85}
\definecolor{mylight}{RGB}{255, 247, 247}
\definecolor{myhighlight}{RGB}{193,210,240}
\definecolor{newnodepurple}{RGB}{184,170,237}
\definecolor{longedgered}{RGB}{192,0,0}
\definecolor{shortedgegreen}{RGB}{0,176,80}
\definecolor{verylightgray}{RGB}{240,240,240} 
\usepackage{pifont}
\newcommand{\cmark}{\textcolor{teal}{\ding{51}}} 
\newcommand{\xmark}{\textcolor{red}{\ding{55}}} 

\begin{document}

\title{Imaginative World Modeling with Scene Graphs via VLM \\ for Object Navigation}


\markboth{IEEE Robotics and Automation Letters. Preprint Version. Accepted August 21, 2026}
{Yue Hu \MakeLowercase{\textit{et al.}}: Imaginative World Modeling with Scene Graphs via VLM for Object Navigation} 

\author{Yue Hu, Member, IEEE, Junzhe Wu, Ruihan Xu, Avery Xi, Hang Liu, \\ 
Henry X. Liu, Senior Member, IEEE, Ram Vasudevan, Maani Ghaffari, Member, IEEE
\thanks{Manuscript received: April 20, 2026; Revised July 14, 2026; Accepted August 21, 2026.}\IEEEcompsocitemizethanks{\scriptsize{This paper was recommended for publication by Editor Pascal Vasseur upon evaluation of the Associate Editor and Reviewers' comments. \IEEEcompsocthanksitem Yue Hu, Junzhe Wu, Ruihan Xu, Hang Liu, Avery Xi, Henry X. Liu, Ram Vasudevan, Maani Ghaffari are with University of Michigan, Ann Arbor, US. Corresponding author is Yue Hu E-mail: huyu@umich.edu. \protect
}}
\thanks{Digital Object Identifier (DOI): see top of this page.}
}


\maketitle


\begin{abstract}

Semantic navigation requires an agent to navigate toward a specified target in an unseen environment. Employing an imaginative navigation strategy that predicts future scenes before taking action, can empower the agent to find target faster. Inspired by this idea, we propose SGImagineNav, a novel imaginative navigation framework that leverages symbolic world modeling to proactively build a global environmental representation. SGImagineNav maintains an evolving hierarchical scene graph and uses large vision language models to predict and explore unseen parts of the environment. While existing methods solely relying on past observations, this imaginative scene graph provides richer semantic context, enabling the agent to proactively estimate target locations. Building upon this, SGImagineNav adopts an adaptive navigation strategy that exploits semantic shortcuts when promising and explores unknown areas otherwise to gather additional context. This strategy continuously expands the known environment and accumulates valuable semantic contexts, ultimately guiding the agent toward the target. SGImagineNav is evaluated in both real-world scenarios and simulation benchmarks. SGImagineNav consistently outperforms previous methods, improving the success rate to \textbf{65.4\%} and \textbf{66.8\%} on HM3D and HSSD, and demonstrating cross-floor and cross-room navigation in real-world environments. All source code is open-sourced at \url{https://github.com/UMich-CURLY/SGImagineNav}.


\end{abstract}

\vspace{-4mm}
\section{Introduction}


Semantic navigation empowers agents to follow open-ended language instructions to locate targets in a new environment. As a foundational capability of embodied AI, it paves the way for agents to effectively interact with and manipulate real-world entities. This functionality has broad applicability, including household assistance, warehouse automation, and outdoor search-and-rescue operations. To enable semantic navigation, a common strategy~\cite{procthor, ramrakhya2023pirlnav,yin2024sgnav, yokoyama2024vlfm, long2024instructnav, zhou2023esc,chaplot2020object} is to construct a world model and apply semantic reasoning to estimate probable target locations. Depending on the world representation, existing navigation approaches can be broadly categorized into three types. First, image-based world models use vision-language models (VLMs)~\cite{liu2023llava} to directly reason about target locations. Although these models offer rich semantic understanding, they are significantly limited in their 3D spatial reasoning capability. Second, map-based world models~\cite{yokoyama2024vlfm,long2024instructnav,huang2024gamap} build spatial layouts of the environment to support geometric reasoning, though they lack high-level semantic abstraction. Third, symbolic world models~\cite{yin2024sgnav,loo2024open,maggio2024clio} represent the environment using structured scene graphs that capture both spatial and semantic cues and support global reasoning over the target locations. In this work, we focus on this symbolic world modeling.

Despite promising advancements, current navigation approaches remain far from human-level performance. A major limitation lies in their construction of world models solely from past observations. In contrast, humans naturally leverage contextual cues and prior knowledge to anticipate unseen areas and reason about target locations before acting. Environments are not randomly arranged; they are structured by semantic priors. This imaginative navigation strategy can create rich semantic shortcuts to locate targets faster. However, very few existing methods~\cite{zhang2024imagine,zhao2024imaginenav} have explored imaginative navigation strategies and neither of them operates on symbolic world. They either depend on computationally heavy image-based predictions~\cite{zhao2024imaginenav}, lacking global cross-view consistency, or map-based predictions~\cite{zhang2024imagine}, lacking high-level semantic reasoning. Furthermore, they all rely on task-specific training, unable to directly adapt and generalize to new environments.

\begin{figure}[t]
    \centering
    \includegraphics[width=1.0\linewidth]{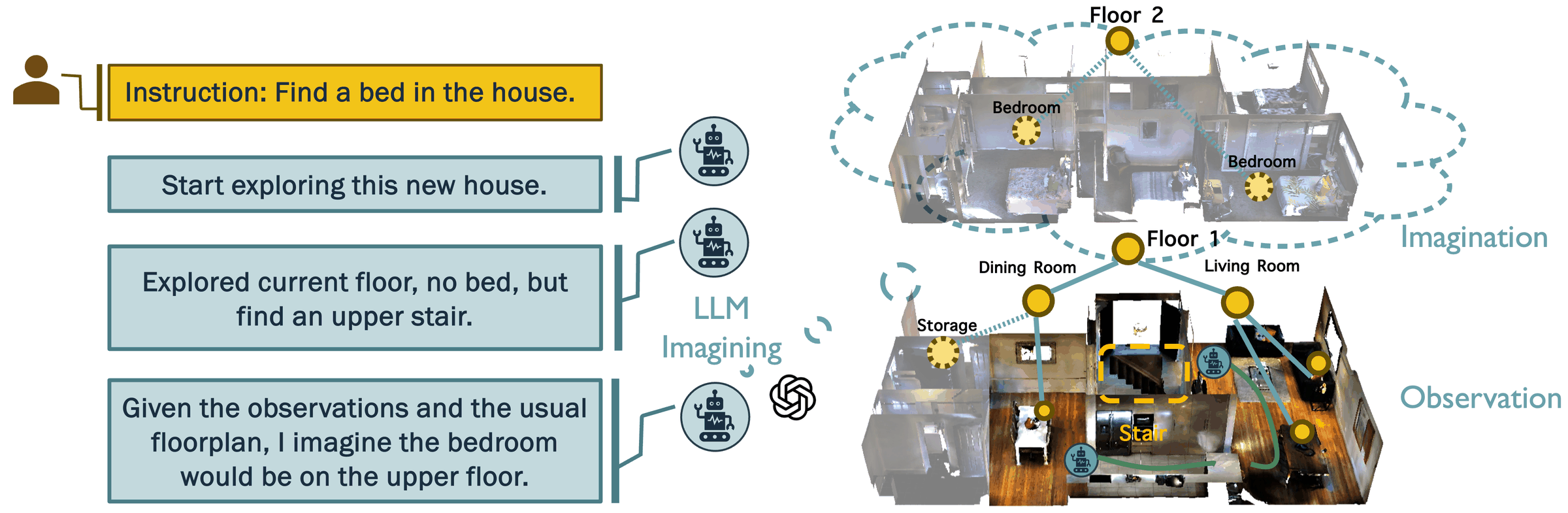}
    \vspace{-4mm}
    \caption{By harnessing the imaginative capabilities of VLMs to anticipate the future in symbolic space with scene graphs, we enable look-ahead navigation, boosting search efficiency.}
    \vspace{-7mm}
    \label{fig:teaser}
\end{figure}

To fill this gap, this paper proposes a novel imaginative navigation framework based on symbolic world model, $\mathtt{SGImagineNav}$. The core idea is to represent the environment using a scene graph and leverage the VLM's prior knowledge about indoor layouts to complete the scene graph in the unexplored areas. This imaginative scene graph supplies rich semantic cues and enables semantic reasoning about target locations before taking action, enabling proactive navigation. Following this paradigm, $\mathtt{SGImagineNav}$ incorporates two key designs. First, a world modeling and imagination module that constructs a proactive hierarchical scene graph to structurally represent the environment. This graph captures semantic entities such as objects, regions, and floors, providing rich contextual information. When searching for a bed, detecting stairs on the main floor enables the agent to infer that a bedroom is located upstairs and to navigate accordingly. Second, a target information gain estimator combines semantic and geometric cues to guide navigation decisions. Rather than randomly exploring the open space, the agent moves toward semantically likely regions when semantic cues are promising; otherwise, it explores larger unknown regions to gather more context.

$\mathtt{SGImagineNav}$ has three distinct advantages. First, it introduces symbolic world modeling with hierarchical structure, allowing for seamless extension to multi-floor environments, unlike prior symbolic methods~\cite{yin2024sgnav,loo2024open}, which are typically constrained to a single floor. Second, by leveraging VLM-powered imagination, $\mathtt{SGImagineNav}$ can proactively complete unknown regions ahead of time, enabling anticipatory navigation, whereas earlier symbolic approaches are limited to past observations~\cite{maggio2024clio}. Third, $\mathtt{SGImagineNav}$ balances broad coverage and target-focused search, while previous approaches either waste steps in uninformative semantic anchors exploitation~\cite{yokoyama2024vlfm,long2024instructnav} or random unknown regions exploration~\cite{ho_kim2024mapex}.


\begin{figure*}[!t]
\centering
\centering{\includegraphics[width=0.99\linewidth]{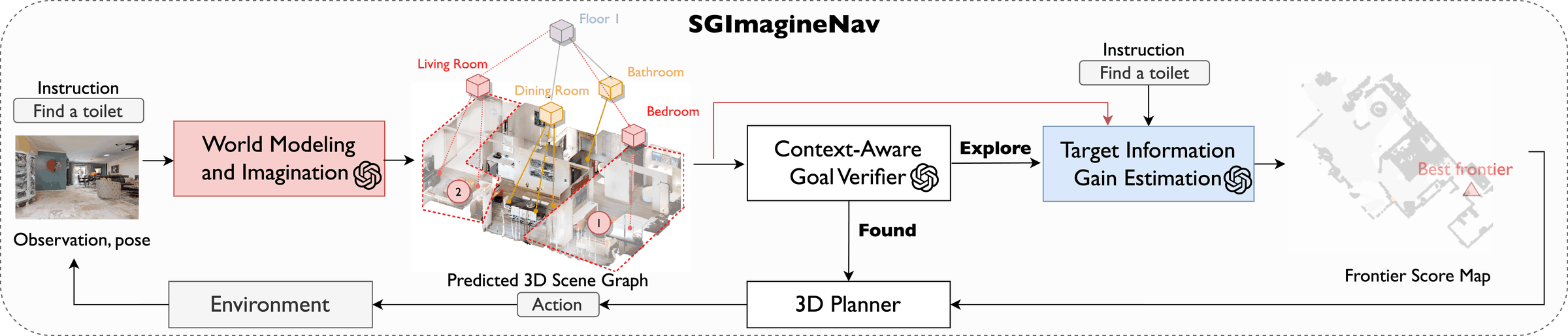}}
\caption{Overview of $\mathtt{SGImagineNav}$. The $\mathtt{SGImagineNav}$ enhances the frontier-based navigation strategy with two key designs: i) world imagination that looks ahead before acting, enabling proactive and informed exploration; ii) target information gain that balances exploitation and exploration gains to guide frontier selection, enabling target-focused search.}
\label{fig:overview}
\vspace{-6mm}
\end{figure*}

To validate the effectiveness of $\mathtt{SGImagineNav}$, we conduct three key evaluations. First, we benchmark $\mathtt{SGImagineNav}$ against prior state-of-the-art methods across two simulation environments: HM3D~\cite{ramakrishnan2021habitat} and HSSD~\cite{khanna2023hssd}. Our approach improves zero-shot SOTA to \textbf{65.4}\% and \textbf{66.8}\% success rate. Notably, it surpasses the previous training-required imaginative navigation SOTA~\cite{zhang2024imagine} by 5.2\%, highlighting its superior capability. Second, we evaluate $\mathtt{SGImagineNav}$ in representative real-world scenarios. The agent successfully navigates across rooms and floors, demonstrating strong generalization capability in complex, open-bounded 3D physical spaces. Third, we conduct ablation studies on the two core components. Results show world modeling and information gain contribute substantial improvements of 6.75\% and 2.75\%, further validating the effectiveness of world modeling and navigation strategy.

To summarize, the primary contributions of this paper are:
    
$\bullet$ $\mathtt{SGImagineNav}$, a novel proactive navigation framework that imagines future world states before physical movement, leading to more informed and efficient planning.

$\bullet$ A symbolic world model grounded in scene graphs and leverage LLMs and VLMs to perform prediction within this symbolic space. 
By harnessing their commonsense knowledge and strong reasoning capabilities, the proposed approach enables the estimation of plausible future world states.

$\bullet$ An experimental validation of $\mathtt{SGImagineNav}$ through comprehensive experiments across diverse benchmarks, showing improvements in success rate and path efficiency.
\section{Related Work}

\textbf{Embodied agent navigation.} Embodied agent navigation involves following a given natural-language instructions and navigating to the specified target. 
The mainstream methods can be divided into three categories: i) training-based methods, such as imitation learning~\cite{ ramrakhya2023pirlnav} and reinforcement learning (RL)~\cite{procthor}, which rely on extensive task-specific training data~~\cite{mousavian2019visual, yang2018visual, majumdar2022zson}.
While these methods can achieve strong performance in trained environments, they often suffer from poor generalization due to domain gaps and limited training diversity, and ii) zero-shot methods, which leverage pre-trained general-purpose models, such as CLIP~\cite{khandelwal2022simple}, LLMs~\cite{openai2023gpt4,team2024gemini}, and VLMs~\cite{liu2023llava}, to perform navigation without additional task-specific training \citep{ zhou2023esc,long2024instructnav,huang2024gamap,yin2024sgnav}; 
and iii) vision-language-action models~\cite{UniNavid:RSS25} that train end-to-end foundation models for direct action prediction from images. Although flexible and adaptable to novel environments, these methods often struggle with geometric reasoning and memory constraints, limiting their effectiveness in long-horizon tasks. Our method adopts a zero-shot strategy by integrating LLMs into a modular system, combining strong generalization and geometric capabilities with optimized task execution and navigation efficiency.

\textbf{World modeling for navigation.} World models~\cite{NEURIPS2018_2de5d166} equip agents with an understanding of their environment, enabling reasoning about target locations and effective navigation. Common representations of world models include images, maps, and symbolic scene graphs. Traditionally, these models are constructed solely from past observations, resulting in an incomplete and partial understanding of the environment. To address this limitation, recent work augments world models with predictive capability, thereby enhancing their ability to anticipate and represent unseen regions. Two primary directions are emerging: i) diffusion-based generative image models~\cite{zhao2024imaginenav}, which generate future visual perspectives; and ii) generative map-based methods~\cite{zhang2024imagine}, which complete the map in unexplored areas. However, both approaches suffer from two key limitations: i) they rely on task-specific training, which restricts their generalizability across diverse environments; and ii) image and map representations often lack the structural richness needed for deep semantic reasoning. Here, we introduce a zero-shot, symbolic approach to imaginative world modeling. It is achieved by using commonsense knowledge from LLMs to predict scene graphs in unseen regions, supporting proactive decision-making for navigation.

\vspace{-2mm}
\section{Imaginative Scene Graph-Based Proactive Navigation Framework}
\vspace{-2mm}

Here, we propose $\mathtt{SGImagineNav}$, a novel proactive navigation framework based on a symbolic world model. First, we achieve a predictive world model through an imaginative scene graph, which extracts rich semantic context from observations and predicts unobserved areas, resulting in a more comprehensive estimation that closely reflects the true environment state. Second, we obtain the exploitation gain and exploration gain through VLM reasoning and accumulated raycasting, and manage the trade-off with a fallback mechanism to switch between exploitation and exploration guidance, resulting in a more balanced search.

\begin{figure*}[!t]
\centering
    \includegraphics[width=0.99\linewidth]{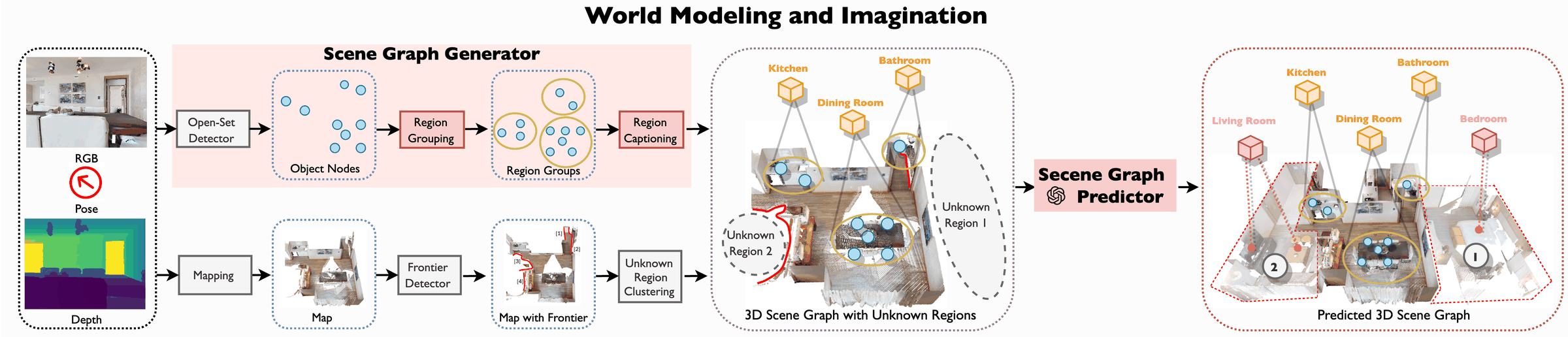}
  \vspace{-1mm}
  \caption{World modeling and imagination module builds a global environmental representation with a hierarchical scene graph and leverages VLM to predict unseen parts, offering a more comprehensive environment state for target search.}
  \label{fig:worldmodel}
\vspace{-6mm}
\end{figure*}

Overall, $\mathtt{SGImagineNav}$ consists of five main modules: a world modeling and imagination module, a context-aware goal verifier, a frontier detector, a target information gain estimator, and a 3D planner. At the core, the world modeling and imagination module constructs a 3D hierarchical scene graph to represent the environment. The context-aware goal verifier checks whether the specified goal is present within the environment. If the goal is found, the 3D planner generates an action plan to navigate toward it. If not, the system switches to a frontier-based exploration strategy. In this case, the target information gain estimator evaluates the detected frontiers~\cite{yamauchi1997frontier} and selects the most informative one. The highest-scoring frontier is set as an intermediate goal and passed to the planner. Through iterative execution of actions and interaction with the environment, the system uncovers previously unknown areas, ultimately guiding the agent toward the target.

\vspace{-2mm}
\subsection{World Modeling and Imagination}\label{sec:construction} 

The world modeling and imagination module constructs a comprehensive symbolic representation of the environment state in the form of a scene graph. Unlike prior work that builds environment maps solely from observed data~\cite{yin2024sgnav, yokoyama2024vlfm}, our approach incorporates an imagination mechanism that predicts unobserved regions, yielding a more complete and informative world model. This comprehensive symbolic abstraction allows the agent to reason beyond current observations and proactively select long-term navigation goals, thereby guiding efficient exploration toward the target object category.


\noindent\textbf{Symbolic world modeling.} Symbolic world modeling extracts the hierarchical semantic entities from the observations and constructs a 3D scene graph. The intuition is that capturing rich semantic cues and structuring them symbolically enables more advanced, context-aware reasoning, ultimately enhancing the accuracy of target prediction. For instance, a bed is typically associated with a bedroom and often found on upper floors.

The hierarchical scene graph consists of nodes and edges organized across three semantic levels: objects, regions, and floors. This multi-level structure effectively captures the semantic organization of typical indoor environments. Nodes represent entities at each level, while edges encode the relationships between them, preserving the hierarchical structure. The graph is built in a bottom-up process, consisting of three steps: i) an open-vocabulary detector identifies objects from visual observations; ii) a region grouper clusters objects based on geometric proximity to define spatial regions, and a region captioner assigns high-level semantic labels to each region, creating region nodes; and iii) a new floor node is initialized whenever the agent transitions to a new floor, typically via stairs. Note that we focus on the topological relationships between hierarchical nodes, rather than computationally expensive fully-connected semantic relationships. As a result, the hierarchical scene graph adopts a tree structure, where edges only connect adjacent levels, objects to regions, and regions to floors. 

Formally, the hierarchical scene graph at timestep $t$, denoted as $g_t = \{\mathcal{V}_t, \mathcal{E}_t\}$, consists of a set of nodes $\mathcal{V}_t$ and edges $\mathcal{E}_t$. Each node is represented by its spatial location, semantic category, and visual feature vector. At each timestep, the world modeling module updates the previous scene graph $g_{t-1}$ with the new observation $o_t$ to produce the current scene graph,
\begin{align*}
    g_t = \phi(g_{t-1},o_t),
\end{align*}
where $\phi(\cdot)$ denotes the scene graph generation function. By integrating information from all past observations, the updated hierarchical scene graph not only reflects the current scene but also maintains temporal consistency, resulting in a comprehensive global environment state. 


\noindent\textbf{Symbolic world imagination.} To further enhance the estimated environment state and make it closer to the actual state, the symbolic world imagination module predicts the unobserved regions and completes the partial scene graph. This allows for proactive target navigation using a more comprehensive environmental representation. To achieve this, it leverages VLM to reason over contextual information in the observed scene graph, incorporating commonsense knowledge about typical indoor layouts to infer the semantics of unobserved areas. The core intuition for the prediction is that environments are inherently structured, for instance, bedrooms are typically near bathrooms, and kitchens are often adjacent to dining rooms. VLMs possess this commonsense knowledge and strong generative reasoning capabilities, enabling them to predict unobserved areas based on surrounding context. Moreover, the symbolic scene graph, easily expressed in a textual format, is naturally compatible with VLMs. The VLM completes the unknown regions in text, which is then transformed into a structured scene graph format. 

\begin{figure*}[!t]
\centering
\includegraphics[width=0.99\linewidth]{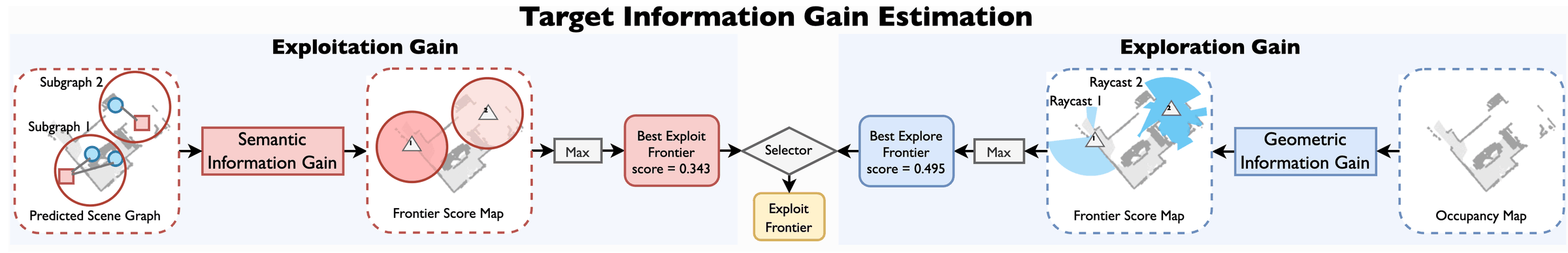}
\vspace{-2mm}
\caption{Target information gain estimation module considers both exploration and exploitation gains to select frontiers. It encourages the system to prioritize regions that are both likely to contain relevant targets and rich in unexplored areas, enabling more informed and efficient target search.}
\label{fig:informationgain}
\vspace{-6mm}
\end{figure*}

Scene graph imagination occurs in two main steps. First, unknown regions within the constructed occupancy map are identified. The centers of these unknown regions, along with the spatial layout of observed scene graph nodes, are projected onto a bird’s-eye view image $I_t$. This explicit localization of unknown areas helps ground the observed scene graph context and its spatial surroundings, allowing the language model to reference typical indoor layouts and reason about the unobserved regions. Second, a VLM is prompted with this spatial context to infer semantic concepts for the unobserved regions. Formally, the completed scene graph $\hat{g}_t$ is defined as,
\begin{align*}
    \hat{g}_t = \psi(g_{t},I_t),
\end{align*}
where $\psi(\cdot)$ is the scene graph prediction function powered by VLM. This completed scene graph $\hat{g}_t$ is closer to the actual environment state $\bar{g}$ compared to the initial scene graph $g_t$. The results in Tab.~\ref{tab:abl_prediction} demonstrate the semantic cost function, which is implemented using scene graph precision and recall metrics, supporting this claim.

The scene graph is updated as new observations arrive, with its two stages running at different rates. Generation is synchronous, updating bottom-up on every observation: i) \emph{Object level}: the detector runs on the new frame and object association creates or updates object nodes; ii) \emph{Region level}: new object nodes trigger region grouping and captioning; and iii) \emph{Floor level}: a floor node is added when the agent traverses a newly detected stair into an unvisited floor. Prediction is asynchronous and runs at a lower rate to save computation, triggered every $20$ timesteps or earlier when a region node changes its label or a new region appears.

\vspace{-2mm}
\subsection{Context-Aware Goal Verifier}\label{sec:Verifier}

Leveraging the completed hierarchical scene graph, our context-aware goal verifier determines if a detected object aligns with the navigation target $q$. Rather than relying solely on object appearance or category prediction from the detector, the verifier incorporates contextual information, such as nearby objects and room types, to assess the likelihood of a match. This is achieved by querying a VLM with an image containing the goal. The contextual image helps disambiguate visually similar objects and reject false positives.

\vspace{-4mm}
\subsection{Target Information Gain Estimation}\label{sec:navigation}

When the goal is not found,  
the system switches to exploration mode
, utilizing target information gain estimation module to identify the most promising frontier for locating the target. We estimate two types of information gain based on the constructed symbolic world model $\hat{g}_t$ to guide the target search. Specifically, the exploitation gain captures semantic correlations between observed entities and the target, while the exploration gain reflects geometric cues that indicate the potential for uncovering large unknown areas. A fallback mechanism is employed to combine these two gains. When a frontier shows a promising semantic shortcut, we prioritize it; however, if the exploration gain suggests that more unknown regions can be uncovered, we trust it to build additional semantic context. By combining both gains, the system balances broad coverage with a targeted search, enhancing the efficiency and effectiveness of target discovery.


\textbf{Target exploitation gain.} Exploitation gain estimates the likelihood of finding the target near each frontier by evaluating the semantic relevance between the observed scene graph and the target. The intuition is that semantic contexts provide shortcuts to find the target faster, e.g, to find a toilet, one can search near a bed or in a bedroom. We use general-purpose models, which possess prior knowledge about the semantic relationships between these entities, to quantify this semantic relevance.


Let $L_t = \{l_t^k\}_{k=1}^{K}$ denote the set of detected frontiers, where each $l_t^k$ represents a spatial location. For each frontier $k$, its exploitation gain is calculated in two steps: i) extracting a local subgraph around each frontier to capture nearby semantic context, resulting in a local node set $\widehat{\mathcal{N}}_t^{k}$; ii) calculating semantic scores between the target category and the hierarchical nodes within the subgraph as, 
\vspace{-1mm}
\begin{align*}
    S_t^{s,k}&=\max_{i\in \widehat{\mathcal{N}}_t^{k}}u_s(v_t^i,q),
\end{align*}
where $q$ denotes the target category and $s(\cdot)$ is the scoring function. Note that, i) the subgraph is pruned geometrically, selecting object nodes within a preset radius of the frontier, along with their connected region nodes; ii) the maximum score is used, assuming that the presence of highly semantically relevant entities signals the need to go toward the frontier; and iii) the scoring function $u_s(\cdot)$ quantifies the semantic relevance of the target based on the observed nodes, which is the key for target-driven navigation. We tried both CLIP similarity and LLM querying. LLM provides commonsense reasoning, while CLIP captures semantic alignments. Both semantic cues can capture semantic shortcuts, guiding the agent to find the target faster.

\textbf{Target exploration gain.} The exploration gain quantifies the amount of unexplored area visible from each frontier, encouraging the agent to discover new regions. This is achieved through deterministic raycasting: based on the occupancy map and frontier positions, we calculate the accumulated exploration gain for the $k$-th frontier as the proportion of visible unknown space, this is, $S_t^{g,k} = \frac{U_t^k}{R}$, where $U_t^k$ is the accumulated visible unknown region size along the path to the frontier, and $R$ is the maximum region size covered by the radius.

Given the exploitation and exploration gains, the frontier is selected using the fallback mechanism as,
\vspace{-2mm}
\begin{equation}
l_t^{k^*} = \arg \max_k 
\begin{cases} 
S_t^{s,k} & \text{if } S_t^{s,k} > \lambda, \\
S_t^{g,k} & \text{otherwise,}
\end{cases}  
\end{equation}
where $\lambda$ is a preset threshold. This ensures that selecting the exploitation frontier when it has a high probability of finding target; otherwise, chose the exploration frontier that can uncover the largest unknown regions. The chosen frontier $l_t^{k^*}$ is then passed to the planner to navigate toward the goal.

\textbf{Comparison with existing information gain for navigation.} Compared to semantic-driven value map-based information gain~\cite{yokoyama2024vlfm,long2024instructnav}, which focuses solely on the most relevant semantic cues, our approach achieves greater efficiency. Since semantic cues cannot be perfectly captured in every individual environment, uncovering more regions helps compensate for any inaccuracies in the semantic-based exploitation strategy. Compared to exploration information gain-only methods~\cite{ho_kim2024mapex}, which aim for complete map coverage, our method avoids redundant steps by excluding unnecessary areas.

\vspace{-4mm}
\subsection{3D Planner}\label{sec:planner}
\vspace{-1mm}

Given the verified goal position or the selected frontier, the planner follows previous works~\cite{yin2024sgnav, yokoyama2024vlfm} and uses the Fast Marching Method (FMM)~\cite{sethian1999fast} to generate the next action. FMM computes a feasible path from the agent's current position to a target position on a 2D occupancy map, which restricts planning to a single floor. Since our hierarchical scene graph models the entire multi-floor environment, the agent must also be able to move between floors to search upstairs or downstairs.

We therefore keep the FMM procedure itself unchanged and replace the 2D occupancy map with a 2.5D representation that augments occupancy with a height map and its gradient. A cell is marked traversable if its height gradient is small enough for the agent to climb, so stairways become part of the traversable set rather than obstacles. Planning with FMM over this map yields cross-floor paths while retaining the efficiency and simplicity of 2D planning.

\begin{figure}[ht!]
    \centering
    \includegraphics[width=0.7\linewidth]{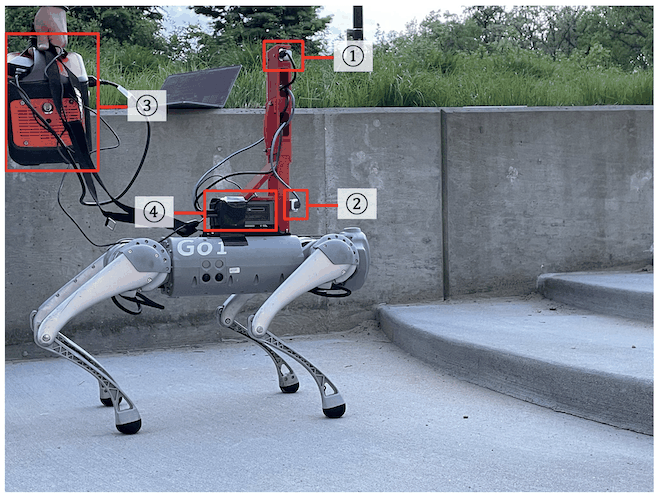}
    \centering
    \vspace{-1mm}
    \caption{Our real-world setup on GO1 legged robot. (1) Intel D435 stereo camera; (2) Intel T265 tracking camera; (3) Portable power bank; (4) NUC mini PC. }
    \label{fig:realworld_setup}
    \vspace{-7mm}
\end{figure}

\vspace{-3mm}
\section{Experimental Results}\label{sec:experiments}
\vspace{-2mm}

\subsection{Experimental setup}
\label{setting}

Here, we detail the implementation details for our navigation pipeline and evaluation setup. All ablation studies are conducted on 400 episodes randomly sampled from HM3D. An episode succeeds if the agent reaches the goal within $0.1$\,m using at most $500$ steps. Baseline results in Table~\ref{tab:main} are quoted from the original papers, which use the same $500$-step budget but a more permissive success distance of $1.0$\,m. The protocols differ only in this threshold, so the comparison is conservative: our numbers are obtained under the stricter criterion.

\textbf{Benchmarks.} We evaluate in both \textbf{real-world} and simulated environments, utilizing our own collected real-world scenarios alongside two widely used public simulation benchmarks, HM3D~\cite{ramakrishnan2021habitat} and HSSD~\cite{khanna2023hssd}. Our real-world dataset encompasses typical, challenging scenarios involving cross-room and cross-floor navigation. HM3D contains 2000 validation episodes in 20 scenes with 6 goal categories. HSSD contains 1248 episodes. 
The real-world setup details are shown in Fig.~\ref{fig:realworld_setup}.



\textbf{Evaluation metrics.} We report three metrics including \emph{success rate (SR)}, \emph{success rate weighted by path length (SPL)}~\cite{spl} and \emph{SoftSPL}. SR is the core metric of object-goal navigation task, representing the success rate of navigation episodes. SPL measures the ability of the agent to find the optimal path. If successful, $\textrm{SPL}=\frac{\textrm{optimal path length}}{\textrm{path length}}$, otherwise $\textrm{SPL}=0$. SoftSPL measures the navigation progress and navigation efficiency of the agent. Higher is better.

\vspace{-3mm}
\subsection{Effectiveness of SGImagineNav}

\textbf{SGImagineNav outperforms previous SOTAs on object-goal navigation benchmarks: HM3D, and HSSD.} Tab.~\ref{tab:main} compares SGImagineNav with several state-of-the-art object navigation methods, including supervised, unsupervised, and zero-shot approaches. SGImagineNav consistently outperforms existing methods across all datasets, improving success rate to \textbf{65.4}\% and \textbf{66.8}\% on HM3D and HSSD. We also notice that: i) SGImagineNav surpasses even supervised methods ProTHOR~\cite{procthor}, demonstrating the effectiveness of our approach. This is because the comprehensive hierarchical scene graph models the entire 3D indoor environment, while previous methods are limited to single-floor search and fundamentally miss objects located on different floors; ii) SGImagineNav also outperforms prior training-required imaginative navigation methods like ImagineNav~\cite{zhao2024imaginenav} and SGM~\cite{zhang2024imagine} by \textbf{12.4}\% and \textbf{5.2}\% on SR, underscoring the effectiveness of symbolic representation in anticipating the future. This is because previous pixel-level map prediction~\cite{zhang2024imagine} lacks semantic reasoning, and image-based imagination~\cite{zhao2024imaginenav} lacks the temporal cues and global reasoning needed for effective target searching.

\begin{table}[!t]
    \centering
    \setlength\tabcolsep{3pt}
    \caption{SGImagineNav outperforms previous SOTAs across different metrics on object-goal navigation benchmark HM3D~\cite{ramakrishnan2021habitat} and HSSD~\cite{khanna2023hssd}. The UnS, ZS, and MF denote unsupervised, zero-shot, and multi-floor, respectively.}
    \vspace{-2mm}
    {
    \begin{tabular}{lcccccccccc}
        \toprule
        \multirow{2}*{\textbf{Method}} & \multirow{2}*{\textbf{UnS}} & \multirow{2}*{\textbf{ZS}}  & \multirow{2}*{\textbf{MF}} &
        \multicolumn{2}{c}{\textbf{HM3D}} & \multicolumn{2}{c}{\textbf{HSSD}} \\
        \cmidrule(lr){5-6} \cmidrule(lr){7-8} \cmidrule(lr){9-10}
        & & && SR$\uparrow$ & SPL$\uparrow$ & SR$\uparrow$ & SPL$\uparrow$ \\
        \midrule
        ImagineNav~\cite{zhao2024imaginenav} & \xmark & \xmark & \xmark & 53.0 & 23.8 & 51.0  & 24.9 \\
        ProcTHOR~\cite{procthor} & \xmark & \xmark & \xmark & 54.4 & \textbf{31.8}  & -- & -- \\
        SGM~\cite{zhang2024imagine} & \xmark & \xmark & \xmark & 60.2 &  30.8  & -- & -- \\
        \midrule
        ProcTHOR-ZS~\cite{procthor} & \xmark & \cmark & \xmark & 13.2 & 7.7 & -- & -- \\
        ZSON~\cite{majumdar2022zson} & \xmark & \cmark & \xmark& 25.5 & 12.6 & -- & -- \\
        \midrule
        ESC~\cite{zhou2023esc} & \cmark & \cmark & \xmark & 39.2 & 22.3  & 38.1 & 22.2 \\
        VoroNav~\cite{wu2024voronav} & \cmark & \cmark & \xmark & 42.0 & 26.0  & 41.0 & 23.2 \\
        L3MVN~\cite{yu2023l3mvn} & \cmark & \cmark & \xmark & 48.7 & 23.0  & 41.2 & 22.5 \\
        VLFM~\cite{yokoyama2024vlfm} & \cmark & \cmark & \xmark & 52.4 & 30.3   & -- & -- \\
        OpenFMNav~\cite{kuang2024openfmnav} & \cmark & \cmark & \xmark & 52.5 & 24.1  & -- & -- \\
        SG-Nav~\cite{yin2024sgnav} & \cmark & \cmark & \xmark & 54.0 & 24.9 & -- & -- \\
        InstructNav~\cite{long2024instructnav} & \cmark & \cmark & \xmark & 58.0 & 20.9  & -- & -- \\
        MFNP~\cite{zhang2024multi} & \cmark & \cmark & \cmark & 58.3 & 26.7  & -- & -- \\		\midrule	
        \rowcolor{mylight} \textbf{SGImagineNav} & \cmark & \cmark & \cmark & \textbf{65.4}  &  30.0 &  \textbf{66.8} &  \textbf{30.2} \\ 
        \bottomrule 
    \end{tabular}
    }
    \label{tab:main}
    \vspace{-3mm}
\end{table}

\begin{table}[!t]
    \centering
    \setlength\tabcolsep{3pt}
    \caption{Ablation of three key modules in the proposed proactive navigation system.}
    \vspace{-2mm}
    \begin{tabular}{cccccccc}
    \toprule
    & \multicolumn{2}{c}{\textbf{World Modeling}}  & \multicolumn{2}{c}{\textbf{Information Gain}} & \textbf{Verification} & \multicolumn{2}{c}{\textbf{HM3D}} \\
    \cmidrule(lr){2-3} \cmidrule(lr){4-5}  \cmidrule(lr){6-6} \cmidrule(lr){7-8} 
    & Generation & Prediction & Exploit & Explore   & Context   & SR$\uparrow$         & SPL$\uparrow$        \\\midrule
    a) & \xmark & \xmark & \xmark & \xmark   & \xmark   & 55.75           &   23.69   \\\midrule
    b) &\cmark & \xmark & \cmark & \xmark  & \xmark    & 57.25   &  24.78    \\
    c) &\cmark & \cmark & \cmark & \xmark  & \xmark    & 62.50    & 27.32   \\
    d) &\cmark & \cmark & \cmark & \cmark  & \xmark    & 65.25       &      31.41       \\
    e) &\cmark & \cmark & \cmark & \cmark  & \cmark    &  67.00          &  31.92            \\
    \midrule
    f) &GT & GT & \cmark & \cmark & - & 72.25 & 38.29  \\
            \bottomrule
        \end{tabular}
    \label{tab:module_abl}
    \vspace{-4mm}
\end{table}
     
\begin{table*}[!t]
  \caption{Ablation of scene graph-based exploitation information gain.}
    \vspace{-2mm}
    \begin{subtable}[t]{0.35\textwidth}
    \centering
    \setlength\tabcolsep{2pt}
    \caption{Scene graph prediction.}
    \vspace{-2mm}
    \begin{tabular}{ccccccc}
            \toprule
            \multirow{2}*{\textbf{Pred}} &\multicolumn{2}{c}{\textbf{Nav Metric}} &\multicolumn{4}{c}{\textbf{Scene Graph Metric}} \\
            \cmidrule(lr){2-3} \cmidrule(lr){4-5}  \cmidrule(lr){6-7}  
             & SR & SPL & R@1 & P@1 & R@3 & P@3 \\\midrule
            \xmark & 57.25   &  24.78  & 17.96 & 26.94 & 33.46 & 47.87 \\
            \midrule
            GPT-4o-mini & 65.25 & 31.00 & 23.41 & 26.26 & 39.27 & 44.36  \\
            GPT-4o & 65.25 & 31.40 & 24.09 & 25.78 & 42.05 & 45.32  \\
            \bottomrule
    \end{tabular}
    \label{tab:abl_prediction}
  \end{subtable}
    \begin{subtable}[t]{0.38\textwidth}
    \centering
    \setlength\tabcolsep{2pt}
    \caption{Scene graph hierarchy.}
    \vspace{-2mm}
    \begin{tabular}{cccccc}
    \toprule
    \multicolumn{3}{c}{\textbf{Scene Graph Hierarchy}} & \multicolumn{3}{c}{\textbf{HM3D}} \\
    \cmidrule(lr){1-3} \cmidrule(lr){4-6} 
    Object & Region       & Floor      & SR$\uparrow$         & SPL$\uparrow$  & SoftSPL$\uparrow$       \\\midrule
    \cmark & \xmark       & \xmark      &    57.75   &  27.34 &    30.12 \\ 
    \cmark & \xmark       & \cmark      &    60.75    & 29.09  & 33.09  \\ 
    \xmark & \cmark       & \cmark      & 65.25       &      31.41      &  34.95 \\
    \cmark & \cmark       & \cmark      &   61.00    & 29.27 &  32.98       \\
            \bottomrule
        \end{tabular}
    \label{tab:abl_hierarchy}
  \end{subtable}
  \begin{subtable}[t]{0.25\textwidth}
    \centering
    \setlength\tabcolsep{2pt}
    \caption{Score function.}
    \vspace{-2mm}
    \begin{tabular}{cccc}
    \toprule
    \multirow{2}*{\textbf{Score}} & \multicolumn{3}{c}{\textbf{HM3D}} \\
     \cmidrule(lr){2-4} 
          & SR$\uparrow$         & SPL$\uparrow$  & SoftSPL$\uparrow$       \\\midrule
    Distance & 55.75 & 23.69  & 26.75\\
    CLIP-V  &    61.25    & 25.05  & 28.78  \\ 
    CLIP-T &   61.50   &  25.85 &  29.70  \\ 
    LLM  &    63.00     &   26.98    &   30.70   \\ 
            \bottomrule
        \end{tabular}
    \label{tab:abl_score}
  \end{subtable}
  \vspace{-4mm}
\end{table*}

\begin{figure*}[!t]
    \includegraphics[width=1.0\textwidth]{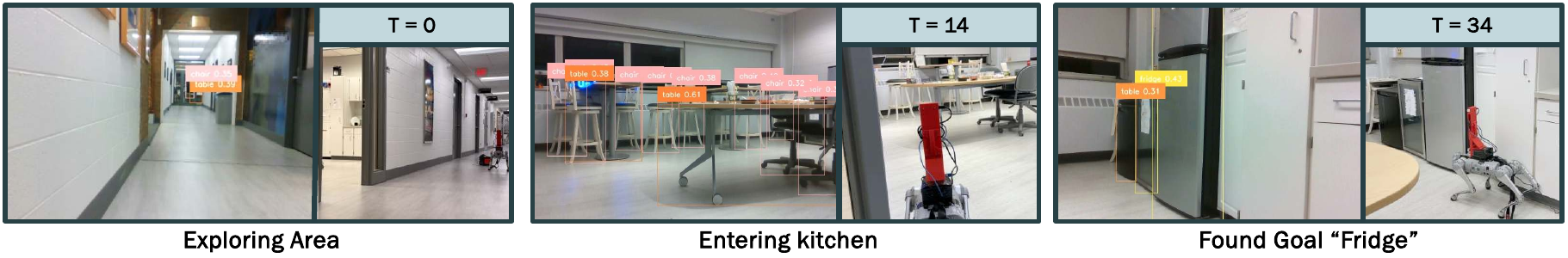}
    \includegraphics[width=1.0\textwidth]{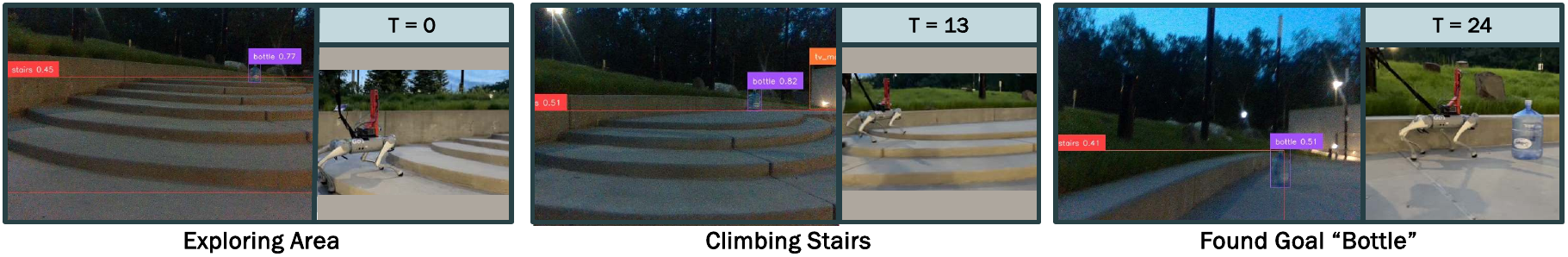}   
    \vspace{-6mm}
    \caption{Successful real-world deployments. The two rows show indoor cross-room navigation and outdoor cross-floor navigation.}
    \label{fig:realworld}
    \vspace{-5mm}
\end{figure*}

\textbf{SGImagineNav deployed on a legged robot successfully searches for targets in real-world cross-floor scenarios.} Fig.~\ref{fig:realworld} shows the real-world navigation process of the robot performing a cross-floor and cross-room target search in the first and second row, respectively. Despite imperfect RGB-D observations and noisy odometry in real-world conditions, SGImagineNav successfully explores the large 3D space, transitions across rooms and floors, and identifies the targets. This real-world deployment highlights the robustness and generalizability of our navigation system.



\vspace{-3mm}
\subsection{Ablation Study}
\vspace{-2mm}

\noindent\textbf{Effectiveness of world modeling, information gain and goal verification.} Tab.~\ref{tab:module_abl} evaluates the effectiveness of our three key modules. The vanilla baseline a) relies on a geometric-only, greedy nearest-frontier selection strategy. 
Each proposed module substantially improves success rate and path efficiency. Specifically: i) variant c) outperforms b), highlighting the effectiveness of the imagination mechanism. The oracle variant f) achieves a 72.25\% success rate, significantly surpassing the baseline and validating the strength of our proactive navigation strategy, further motivating exploration into imaginative navigation; ii) variant d), which integrates both exploitation and exploration gains, outperforms variants b), emphasizing the importance of balancing semantic and geometric cues for efficient target search; iii) goal verification boosts success rate by leveraging contextual information to reject false positives, improving search accuracy.

\textbf{Effectiveness of scene graph prediction.} Tab.~\ref{tab:abl_prediction} evaluates the impact of world imagination and the influence of different LLMs used as predictors. We observe that GPT-4o-mini and GPT-4o perform comparably, both enhancing scene graph recall and enabling a more comprehensive representation of the environment. This suggests that scene graph prediction is robust to the choice of VLM, with even smaller models effectively capturing global arrangement knowledge. Specifically, we see that: i) enabling prediction improves scene graph recall, leading to a more comprehensive representation of the environment; and ii) incorporating prediction leads to a slight drop in precision, which is expected since predictions based on context are generally less accurate than direct observations. The substantial improvement in recall outweighs this precision loss, ultimately benefiting the final navigation.




\textbf{Effects of hierarchical semantics.} Tab.~\ref{tab:abl_hierarchy} evaluates the effectiveness of the three semantic hierarchies. We see that: i) excluding the floor level leads to the most significant decline in SR and SPL. This is because targets are often located on different floors from the agent, and ignoring this semantic level fundamentally limits navigation performance; ii) region-only outperforms object-only, demonstrating that higher-level semantic information provides more informative cues for target search; and iii) combining both object and region semantics results in inferior performance compared to region-only. 
The reason is that object-level semantics can be redundant and noisy while region-level semantics, formed by grouping object features, are more robust to misdetections and better capture the context of object co-occurrence. This is supported by the evidence that both object- and region-level semantics outperformed prior value map methods~\cite{yokoyama2024vlfm}. See the appendix for detailed examples.



\textbf{Effects of score function.} Tab.~\ref{tab:abl_score} compares the distance-based with three variants of semantic relevance score functions: CLIP-V and CLIP-T, which compute cosine similarity between the node’s visual features or category text and the target text, and LLM, which prompts a language model for relevance estimation. We see that: i) all semantic cues outperform the distance-based method, indicating that semantic information aligns well with indoor layouts and effectively guides the agent, providing a shortcut for target search; and (ii) LLM outperforms both CLIP variants, demonstrating its superior reasoning ability for target location while CLIP models are primarily designed for alignment and not targeted for co-occurrence attributes essential for navigation.

\begin{table}[!ht]
    \centering
    \caption{Scene graph prediction is robust, producing consistent results across repeated inferences.}
    \label{tab:abl_robustness}
    \begin{tabular}{lcccc}
\toprule
& Recall@1 & Precision@1 & Recall@3 & Precision@3 \\
\midrule
Repeat 1 & 22.2 & 18.2 & 39.7 & 33.2 \\
Repeat 2 & 21.8 & 18.1 & 39.4 & 33.1 \\
Repeat 3 & 21.9 & 18.2 & 39.0 & 33.0 \\
\bottomrule
\end{tabular}
\vspace{-3mm}
\end{table}

\begin{table}[h!]
    \centering
    \setlength\tabcolsep{4pt}
    \caption{Hierarchical scene graph maintains strong performance across both simple and complex scenes.}
    \label{tab:abl_scalability}
    \begin{tabular}{lccccc}
    \toprule
    \textbf{Scene} & \textbf{Pred} & \textbf{R@1} & \textbf{R@3} & \textbf{P@1} & \textbf{P@3} \\
    \midrule
    Simple & - & 20.5 & 36.5 & 22.1 & 39.8 \\
    Simple & \cmark & 24.6 (+4.1) & 41.3 (+4.8) & 22.9 (+0.8) & 39.9 (+0.1) \\
    \midrule
    Complex & - & 20.8 & 38.2 & 18.1 & 33.4 \\
    Complex & \cmark & 22.5 (+1.7) & 40.3 (+2.2) & 18.5 (+0.4) & 33.8 (+0.4) \\
    \bottomrule
    \end{tabular}
    \vspace{-5mm}
\end{table}

\textbf{Robustness of scene graph prediction.} Tab.~\ref{tab:abl_robustness} evaluates the robustness of our prediction module. To quantitatively assess consistency, we performed three repeated inferences using identical inputs. The results show: i) a low standard deviation of 0.0513 in confidence scores, indicating high prediction stability; and ii) consistent scene graph quality across runs in terms of both precision and recall. These findings suggest that VLMs possess strong knowledge of global structural arrangements and exhibit stable, reliable performance.

\textbf{Scalability of the hierarchical scene graph.} Tab.~\ref{tab:abl_scalability} assesses scene graph quality across both simple and complex scenes. The Simple and Complex scenarios involve approximately 40 and 90 objects, respectively. Pred indicates whether or not scene graph prediction is used, while R\@N and P\@N represent top-N recall and precision, respectively. We see that: i) the prediction module consistently improves performance across varying scene complexities, even when the number of objects is more than doubled, highlighting its robustness; and ii) the scene graph quality in complex scenes remains comparable to that in simple scenes, demonstrating the scalability of the hierarchical scene graph in handling increased complexity.

\vspace{-4mm}
\subsection{Qualitative Analysis}
\vspace{-1mm}

\begin{figure}[!t]
    \centering{\includegraphics[width=1.0\linewidth]{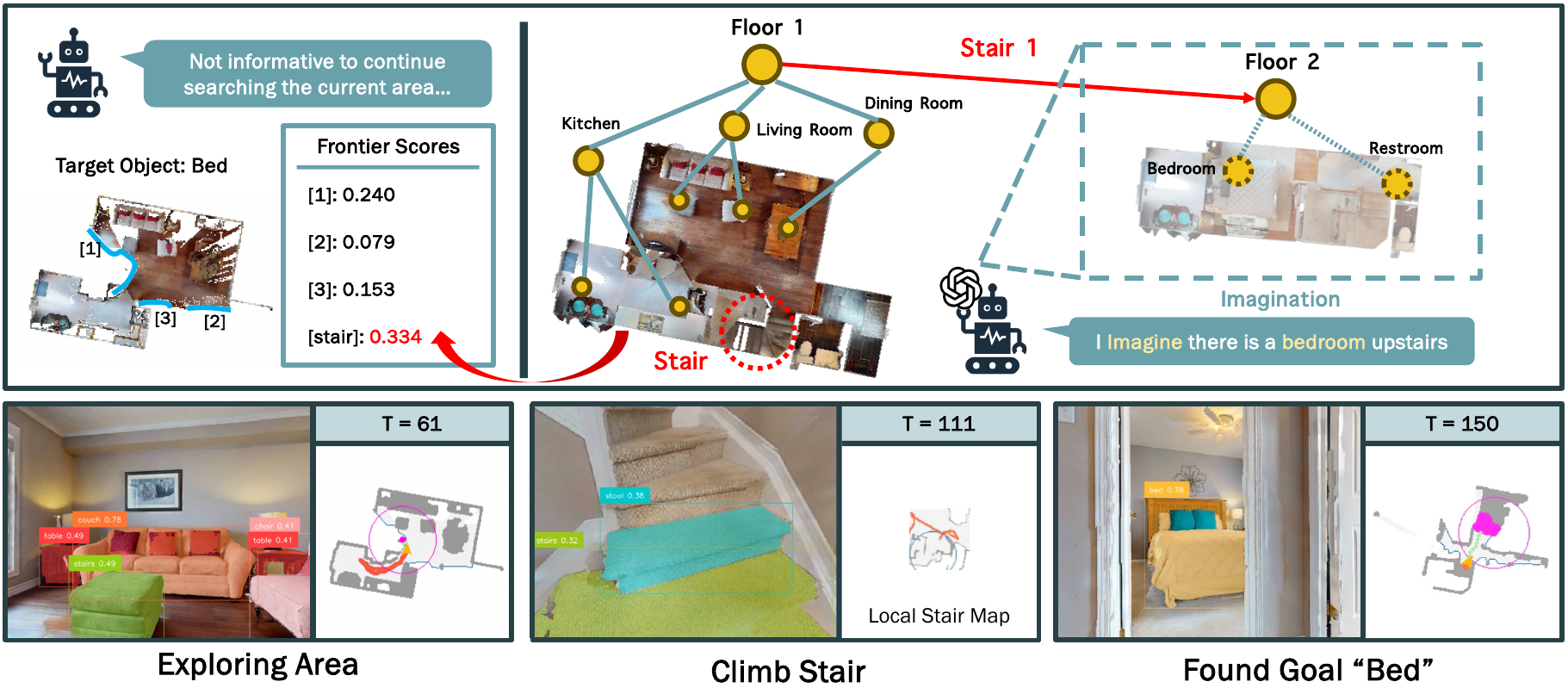}}
    \vspace{-6mm}
    \caption{During navigation, the agent explores the first floor and builds a hierarchical scene graph. After several steps, the imagined scene graph then suggests a potential target located upstairs. As a result, the agent selects the stairway frontier and goes upstairs. Finally, it locates the target.}
    \vspace{-6mm}
    \label{fig:qualitative}
\end{figure}

\textbf{Visualization of navigation process.} Fig.~\ref{fig:qualitative} shows the navigation process. The agent begins by exploring the first floor, gradually building a hierarchical scene graph. When the imagined scene graph identifies a potential target location upstairs, the agent selects the stairway frontier and ascends. Upon reaching the second floor, it successfully locates the target. The imagined hierarchical scene graph offers an anticipatory and structured understanding of the indoor environment, enabling global reasoning about potential target locations. The integrated information gain guides the agent to explore more informative regions with a higher probability of finding the target. Together, these modules empower the agent to efficiently perform cross-floor search and locate targets in large-scale spaces.

\begin{figure}[!t]
    \centering
    \vspace{-4mm}
    \includegraphics[width=0.6\linewidth]{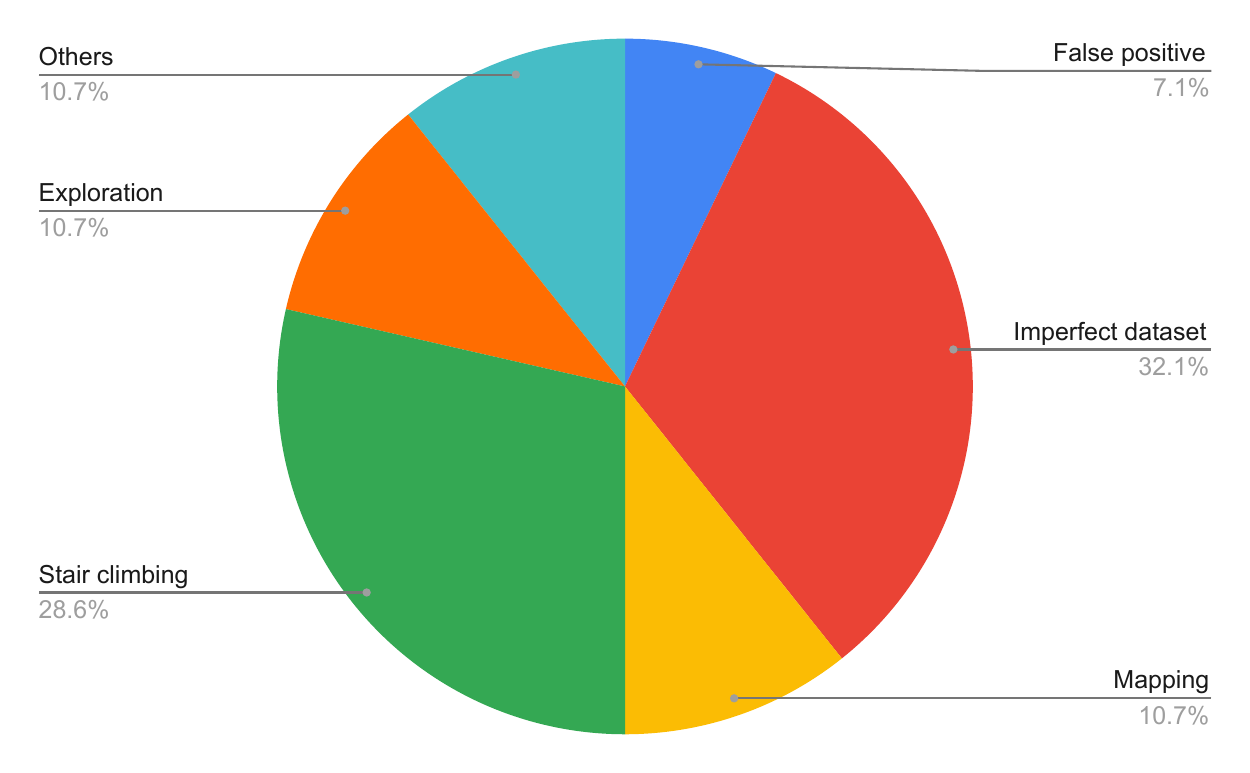}
    \centering
    \vspace{-2mm}
    \caption{Visualization of the error distribution. The top five errors are from i) false positive detections, ii) imperfect dataset, iii) mapping, iv) stair climbing, and v) exploration.}
    \label{fig:error_analysis}
    \vspace{-7mm}
\end{figure}

\vspace{-4mm}
\subsection{Error Analysis}
\vspace{-1mm}




We perform error analysis using the same data split as in the ablation studies. The quantitative error distribution is summarized in Fig.~\ref{fig:error_analysis}. We identify five common error types: i) imperfect or noisy dataset annotations, ii) false positives in perception, iii) mapping errors in perception, iv) planning failures during stair climbing, and v) exploration errors caused by redundant navigation and repetitive exploration strategies. Note that, considering the dataset imperfections, the upper bound performance on HM3D is approximately $88.84\%$. Our method achieves encouraging performance 65.4\%.

\vspace{-4mm}
\section{Conclusion}\label{section:Conclusion}

In this work, we introduce SGImagineNav, a novel proactive navigation system built on symbolic world modeling. SGImagineNav constructs a hierarchical scene graph that captures a wide spectrum of semantic concepts in indoor environments, offering rich contextual cues for target search. Furthermore, its symbolic world imagination mechanism allows the agent to infer unseen areas based on prior knowledge of typical indoor layouts. This comprehensive representation enables proactive search strategies, significantly enhancing the agent's exploration efficiency. Extensive evaluations in both real-world and simulation scenarios demonstrate that SGImagineNav can locate targets across diverse indoor settings, including cross-room and cross-floor environments.

\textbf{Limitations.} SGImagineNav currently lacks sequential planning but can be extended using its imagination mechanism to predict environmental changes and long-term rewards, reducing temporal redundancy and searching the target more quickly.


{\small
\bibliographystyle{unsrt}
\bibliography{ref}

\begin{thebibliography}{10}

\bibitem{procthor}
Matt Deitke, Eli VanderBilt, Alvaro Herrasti, Luca Weihs, Jordi Salvador, Kiana Ehsani, Winson Han, Eric Kolve, Ali Farhadi, Aniruddha Kembhavi, and Roozbeh Mottaghi.
\newblock Procthor: Large-scale embodied ai using procedural generation, 2022.

\bibitem{ramrakhya2023pirlnav}
Ram Ramrakhya, Dhruv Batra, Erik Wijmans, and Abhishek Das.
\newblock Pirlnav: Pretraining with imitation and rl finetuning for objectnav.
\newblock In {\em IEEE/CVF Conference on Computer Vision and Pattern Recognition}, pages 17896--17906, 2023.

\bibitem{yin2024sgnav}
Hang Yin, Xiuwei Xu, Zhenyu Wu, Jie Zhou, and Jiwen Lu.
\newblock Sg-nav: Online 3d scene graph prompting for llm-based zero-shot object navigation.
\newblock {\em arXiv preprint arXiv:2410.08189}, 2024.

\bibitem{yokoyama2024vlfm}
Naoki Yokoyama, Sehoon Ha, Dhruv Batra, Jiuguang Wang, and Bernadette Bucher.
\newblock Vlfm: Vision-language frontier maps for zero-shot semantic navigation.
\newblock In {\em International Conference on Robotics and Automation (ICRA)}, pages 42--48. IEEE, 2024.

\bibitem{long2024instructnav}
Yuxing Long, Wenzhe Cai, Hongcheng Wang, Guanqi Zhan, and Hao Dong.
\newblock Instructnav: Zero-shot system for generic instruction navigation in unexplored environment.
\newblock {\em arXiv preprint arXiv:2406.04882}, 2024.

\bibitem{zhou2023esc}
Kaiwen Zhou, Kaizhi Zheng, Connor Pryor, Yilin Shen, Hongxia Jin, Lise Getoor, and Xin~Eric Wang.
\newblock Esc: Exploration with soft commonsense constraints for zero-shot object navigation.
\newblock In {\em International Conference on Machine Learning}, pages 42829--42842. PMLR, 2023.

\bibitem{chaplot2020object}
Devendra~Singh Chaplot, Dhiraj~Prakashchand Gandhi, Abhinav Gupta, and Russ~R Salakhutdinov.
\newblock Object goal navigation using goal-oriented semantic exploration.
\newblock {\em Advances in Neural Information Processing Systems}, 33:4247--4258, 2020.

\bibitem{liu2023llava}
Haotian Liu, Chunyuan Li, Qingyang Wu, and Yong~Jae Lee.
\newblock Visual instruction tuning, 2023.

\bibitem{huang2024gamap}
Hao Huang, Yu~Hao, Congcong Wen, Anthony Tzes, Yi~Fang, et~al.
\newblock Gamap: Zero-shot object goal navigation with multi-scale geometric-affordance guidance.
\newblock {\em Advances in Neural Information Processing Systems}, 37:39386--39408, 2024.

\bibitem{loo2024open}
Joel Loo, Zhanxin Wu, and David Hsu.
\newblock Open scene graphs for open world object-goal navigation.
\newblock {\em arXiv preprint arXiv:2407.02473}, 2024.

\bibitem{maggio2024clio}
Dominic Maggio, Yun Chang, Nathan Hughes, Matthew Trang, Dan Griffith, Carlyn Dougherty, Eric Cristofalo, Lukas Schmid, and Luca Carlone.
\newblock Clio: Real-time task-driven open-set 3d scene graphs.
\newblock {\em IEEE Robotics and Automation Letters}, 2024.

\bibitem{zhang2024imagine}
Sixian Zhang, Xinyao Yu, Xinhang Song, Xiaohan Wang, and Shuqiang Jiang.
\newblock Imagine before go: Self-supervised generative map for object goal navigation.
\newblock In {\em Computer Vision and Pattern Recognition}, pages 16414--16425, 2024.

\bibitem{zhao2024imaginenav}
Xinxin Zhao, Wenzhe Cai, Likun Tang, and Teng Wang.
\newblock Imaginenav: Prompting vision-language models as embodied navigator through scene imagination.
\newblock {\em arXiv:2410.09874}, 2024.

\bibitem{ho_kim2024mapex}
Cherie Ho, Seungchan Kim, Brady Moon, Aditya Parandekar, Narek Harutyunyan, Chen Wang, Katia Sycara, Graeme Best, and Sebastian Scherer.
\newblock Mapex: Indoor structure exploration with probabilistic information gain from global map predictions.
\newblock {\em arXiv preprint arXiv:2409.15590}, 2024.

\bibitem{ramakrishnan2021habitat}
Santhosh~K Ramakrishnan, Aaron Gokaslan, Erik Wijmans, Oleksandr Maksymets, Alex Clegg, John Turner, Eric Undersander, Wojciech Galuba, Andrew Westbury, Angel~X Chang, et~al.
\newblock Habitat-matterport 3d dataset (hm3d): 1000 large-scale 3d environments for embodied ai.
\newblock {\em arXiv:2109.08238}, 2021.

\bibitem{khanna2023hssd}
Mukul {Khanna*}, Yongsen {Mao*}, Hanxiao Jiang, Sanjay Haresh, Brennan Shacklett, Dhruv Batra, Alexander Clegg, Eric Undersander, Angel~X. Chang, and Manolis Savva.
\newblock {Habitat Synthetic Scenes Dataset (HSSD-200): An Analysis of 3D Scene Scale and Realism Tradeoffs for ObjectGoal Navigation}.
\newblock {\em arXiv}, 2023.

\bibitem{mousavian2019visual}
Arsalan Mousavian, Alexander Toshev, Marek Fi{\v{s}}er, Jana Ko{\v{s}}eck{\'a}, Ayzaan Wahid, and James Davidson.
\newblock Visual representations for semantic target driven navigation.
\newblock In {\em 2019 International Conference on Robotics and Automation (ICRA)}, pages 8846--8852. IEEE, 2019.

\bibitem{yang2018visual}
Wei Yang, Xiaolong Wang, Ali Farhadi, Abhinav Gupta, and Roozbeh Mottaghi.
\newblock Visual semantic navigation using scene priors.
\newblock {\em arXiv preprint arXiv:1810.06543}, 2018.

\bibitem{majumdar2022zson}
Arjun Majumdar, Gunjan Aggarwal, Bhavika Devnani, Judy Hoffman, and Dhruv Batra.
\newblock Zson: Zero-shot object-goal navigation using multimodal goal embeddings.
\newblock {\em Advances in Neural Information Processing Systems}, 35:32340--32352, 2022.

\bibitem{khandelwal2022simple}
Apoorv Khandelwal, Luca Weihs, Roozbeh Mottaghi, and Aniruddha Kembhavi.
\newblock Simple but effective: Clip embeddings for embodied ai.
\newblock In {\em CVPR}, pages 14829--14838, 2022.

\bibitem{openai2023gpt4}
OpenAI.
\newblock Gpt-4 technical report.
\newblock {\em arXiv:2303.08774}, 2023.

\bibitem{team2024gemini}
Gemini Team.
\newblock Gemini 1.5: Unlocking multimodal understanding across millions of tokens of context.
\newblock {\em arXiv:2403.05530}, 2024.

\bibitem{UniNavid:RSS25}
Jiazhao Zhang, Kunyu Wang, Shaoan Wang, Minghan Li, Haoran Liu, Songlin Wei, Zhongyuan Wang, Zhizheng Zhang, and He~Wang.
\newblock Uni-navid: A video-based vision-language-action model for unifying embodied navigation tasks.
\newblock {\em Robotics: Science and Systems}, 2025.

\bibitem{NEURIPS2018_2de5d166}
David Ha and J\"{u}rgen Schmidhuber.
\newblock Recurrent world models facilitate policy evolution.
\newblock In {\em Advances in Neural Information Processing Systems}, volume~31, 2018.

\bibitem{yamauchi1997frontier}
Brian Yamauchi.
\newblock A frontier-based approach for autonomous exploration.
\newblock In {\em International Symposium on Computational Intelligence in Robotics and Automation}. IEEE, 1997.

\bibitem{sethian1999fast}
James~A Sethian.
\newblock Fast marching methods.
\newblock {\em SIAM review}, 41(2):199--235, 1999.

\bibitem{spl}
Peter Anderson, Angel~X. Chang, Devendra~Singh Chaplot, Alexey Dosovitskiy, Saurabh Gupta, Vladlen Koltun, Jana Kosecka, Jitendra Malik, Roozbeh Mottaghi, Manolis Savva, and Amir~R. Zamir.
\newblock On evaluation of embodied navigation agents.
\newblock {\em CoRR}, abs/1807.06757, 2018.

\bibitem{wu2024voronav}
Pengying Wu, Yao Mu, Bingxian Wu, Yi~Hou, Ji~Ma, Shanghang Zhang, and Chang Liu.
\newblock Voronav: Voronoi-based zero-shot object navigation with large language model.
\newblock {\em arXiv preprint arXiv:2401.02695}, 2024.

\bibitem{yu2023l3mvn}
Bangguo Yu, Hamidreza Kasaei, and Ming Cao.
\newblock L3mvn: Leveraging large language models for visual target navigation.
\newblock In {\em 2023 IEEE/RSJ International Conference on Intelligent Robots and Systems (IROS)}, pages 3554--3560. IEEE, 2023.

\bibitem{kuang2024openfmnav}
Yuxuan Kuang, Hai Lin, and Meng Jiang.
\newblock Openfmnav: Towards open-set zero-shot object navigation via vision-language foundation models.
\newblock {\em arXiv preprint arXiv:2402.10670}, 2024.

\bibitem{zhang2024multi}
Lingfeng Zhang, Hao Wang, Erjia Xiao, Xinyao Zhang, Qiang Zhang, Zixuan Jiang, and Renjing Xu.
\newblock Multi-floor zero-shot object navigation policy.
\newblock {\em arXiv preprint arXiv:2409.10906}, 2024.

\end{thebibliography}
}

\newpage
\section{Appendix}
\subsection{Scene graph}

We detail the scene graph imagination module across five key dimensions: i) scene graph representation, ii) scene graph generation and serialization, iii) scene graph prediction and prompting, iv) the updating mechanism, and v) the handling of hallucinated predictions.

\textbf{1. Scene graph representation.} 
We define and represent the scene graph as follows:

\begin{itemize}
\item \textbf{Scene graph definition.} To incorporate rich semantics, we construct a hierarchical scene graph organized across three distinct semantic levels: objects, regions, and floors. Nodes represent entities at each level, while edges encode their spatial and structural relationships to preserve the hierarchy. Note that, our approach focuses on topological relationships between these hierarchical nodes rather than computationally expensive, fully connected semantic relationships. Consequently, the hierarchical scene graph adopts a tree structure where edges only connect adjacent levels (i.e., objects to regions, and regions to floors).
\item \textbf{Input representation for foundation models.} To interface seamlessly with LLMs and VLMs, this structured scene graph is formatted as a dictionary (\texttt{dict}) to serve as context. A concrete example is provided in Tab.~\ref{tab:scene_graph_example_input}.
\end{itemize}


\begin{responselong}[%
   title={Observed scene graph example},%
   label={tab:scene_graph_example_input},%
]
{
  "floors": [
    {
      "id": "0",
      "regions": [
        {
          "id": 4,
          "caption": {
            "bathroom": 0.1810302734375,
            "laundry room": 0.153564453125
          },
          "center": [308.9453430175781, 221.08883666992188],
          "grid_bbox": [[293.8847351074219, 237.10118103027344], [309.3623962402344, 199.11061096191406], [324.0059509277344, 205.0764923095703], [308.5282897949219, 243.0670623779297]],
          "corr_score": [0.2, 0.2],
          "objects": [
            {
              "id": 11,
              "caption": "sink",
              "center": [299, 235],
              "confidence": 0.36839479207992554,
              "corr_score": 0.43603515625
            },
            {
              "id": 68,
              "caption": "cabinet",
              "center": [300, 235],
              "confidence": 0.37238651514053345,
              "corr_score": 0.5390625
            },
            {
              "id": 132,
              "caption": "towel",
              "center": [310, 202],
              "confidence": 0.39796754717826843,
              "corr_score": 0.39208984375
            },
            {
              "id": 158,
              "caption": "bathtub",
              "center": [302, 239],
              "confidence": 0.32076823711395264,
              "corr_score": 0.465576171875
            }
          ]
        },
        ...
      ]
    }
  ]
}

\end{responselong}

\begin{responselong}[%
   title={Predicted scene graph example},%
   label={tab:scene_graph_example_output},%
]
{
  "floors": [
    {
      "id": "0",
      "regions": [
        {
          "id": "0.0",
          "caption": "bedroom",
          "center": [302, 125],
          "corr_score": 0,
          "predicted": true,
          "objects": [
            {
              "id": "0.0.0",
              "caption": "bed",
              "center": [301, 124],
              "confidence": 0.85,
              "corr_score": 0.5
            },
            {
              "id": "0.0.1",
              "caption": "cabinet",
              "center": [303, 126],
              "confidence": 0.75,
              "corr_score": 0.4
            },
            {
              "id": "0.0.2",
              "caption": "tv",
              "center": [304, 127],
              "confidence": 0.7,
              "corr_score": 0.9
            }
          ]
        },
        ...
      ]
    }
  ]
}

\end{responselong}

\textbf{2. Scene graph generation and prompts.} This hierarchical scene graph is built in a bottom-up process, consisting of three steps: i) an open-vocabulary detector identifies objects from visual observations and associate them with historical objects, creating object nodes; ii) a region grouper clusters objects based on geometric proximity to define spatial regions, and a region captioner assigns high-level semantic labels to each region, creating region nodes; and iii) a new floor node is initialized whenever the agent transitions to a new floor, typically via stairs.

\begin{figure*}[!t]
\centering
\includegraphics[width=0.24\textwidth]{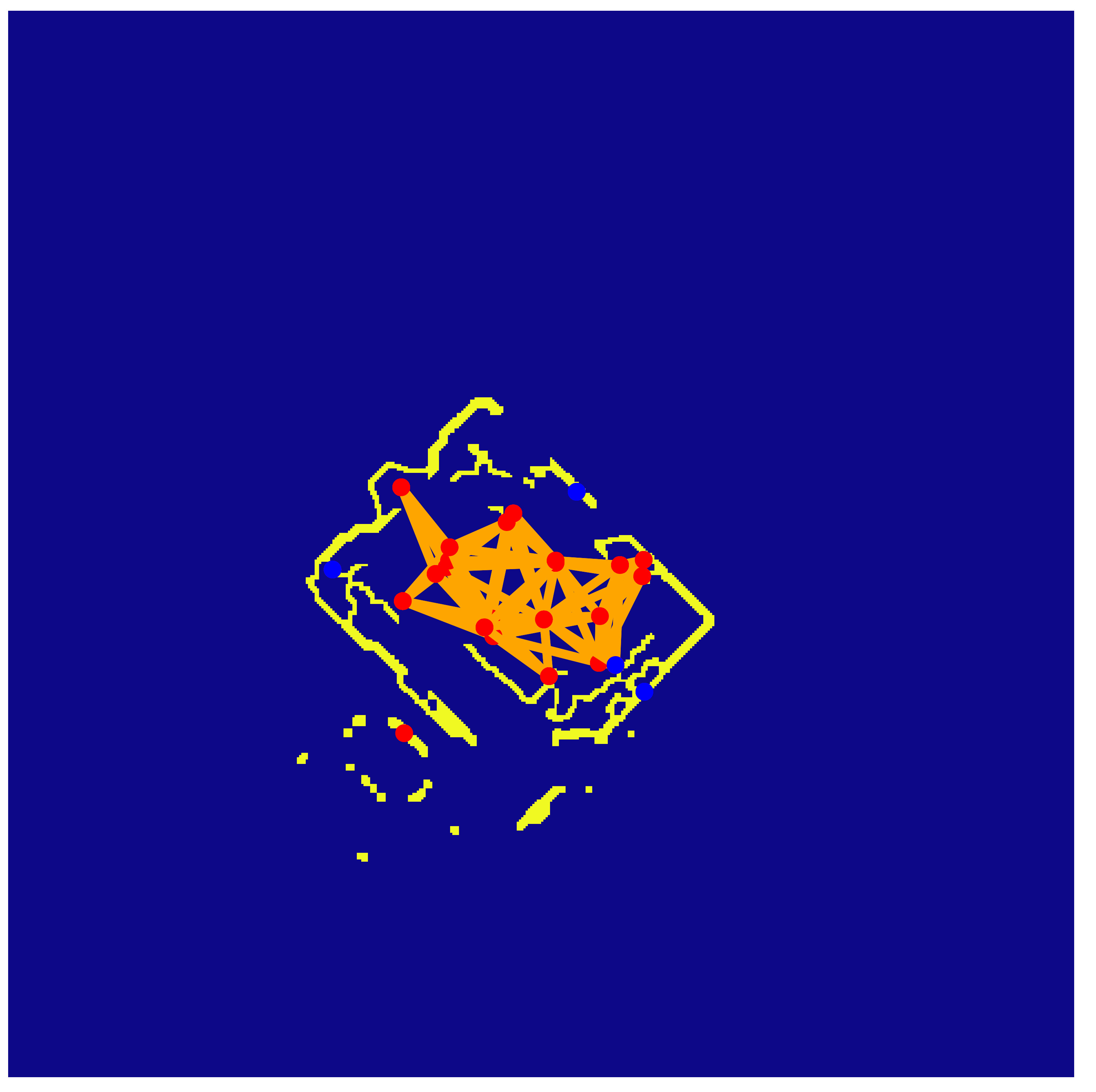}
\includegraphics[width=0.24\textwidth]{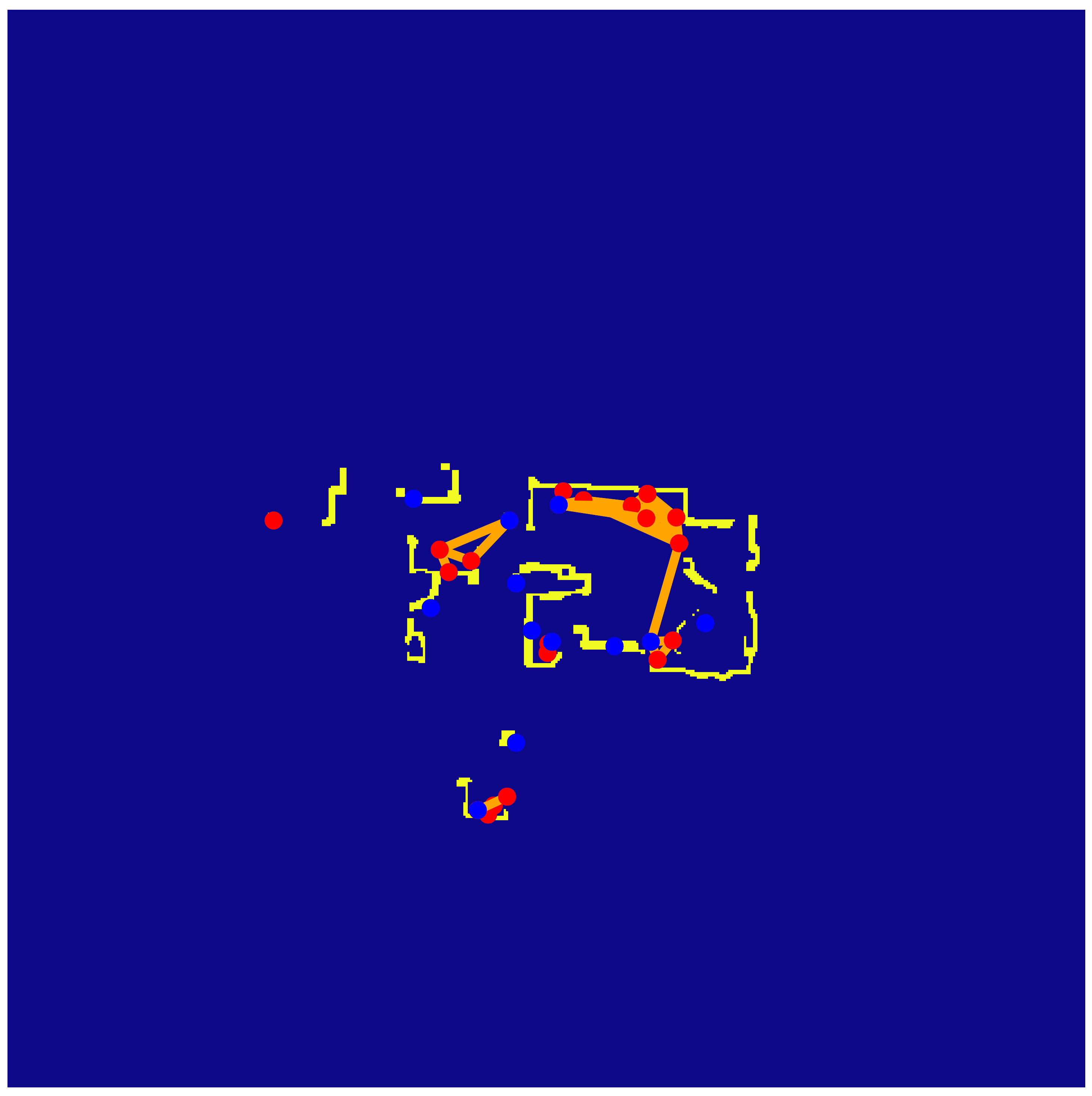}
\includegraphics[width=0.24\textwidth]{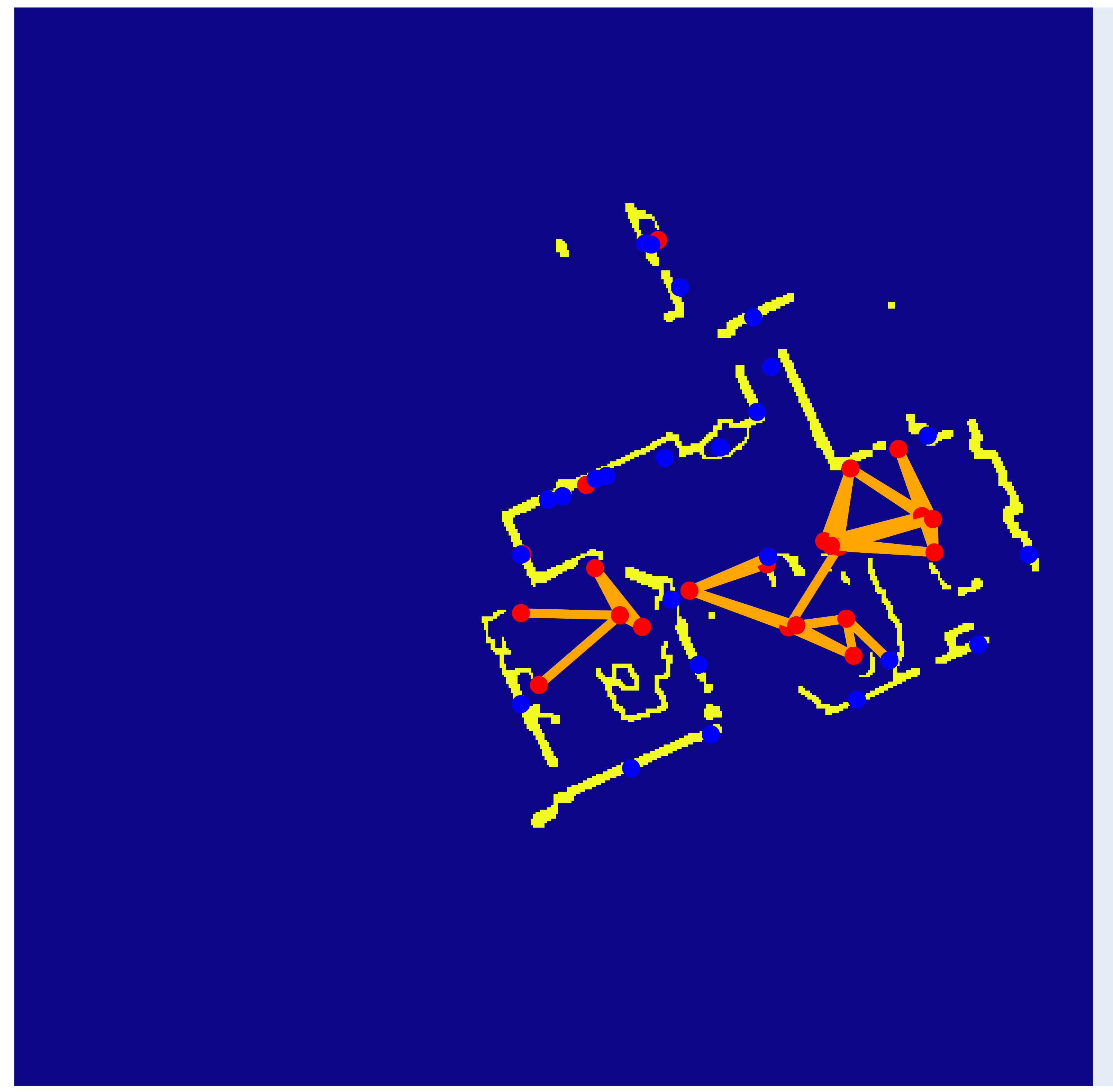}
\includegraphics[width=0.24\textwidth]{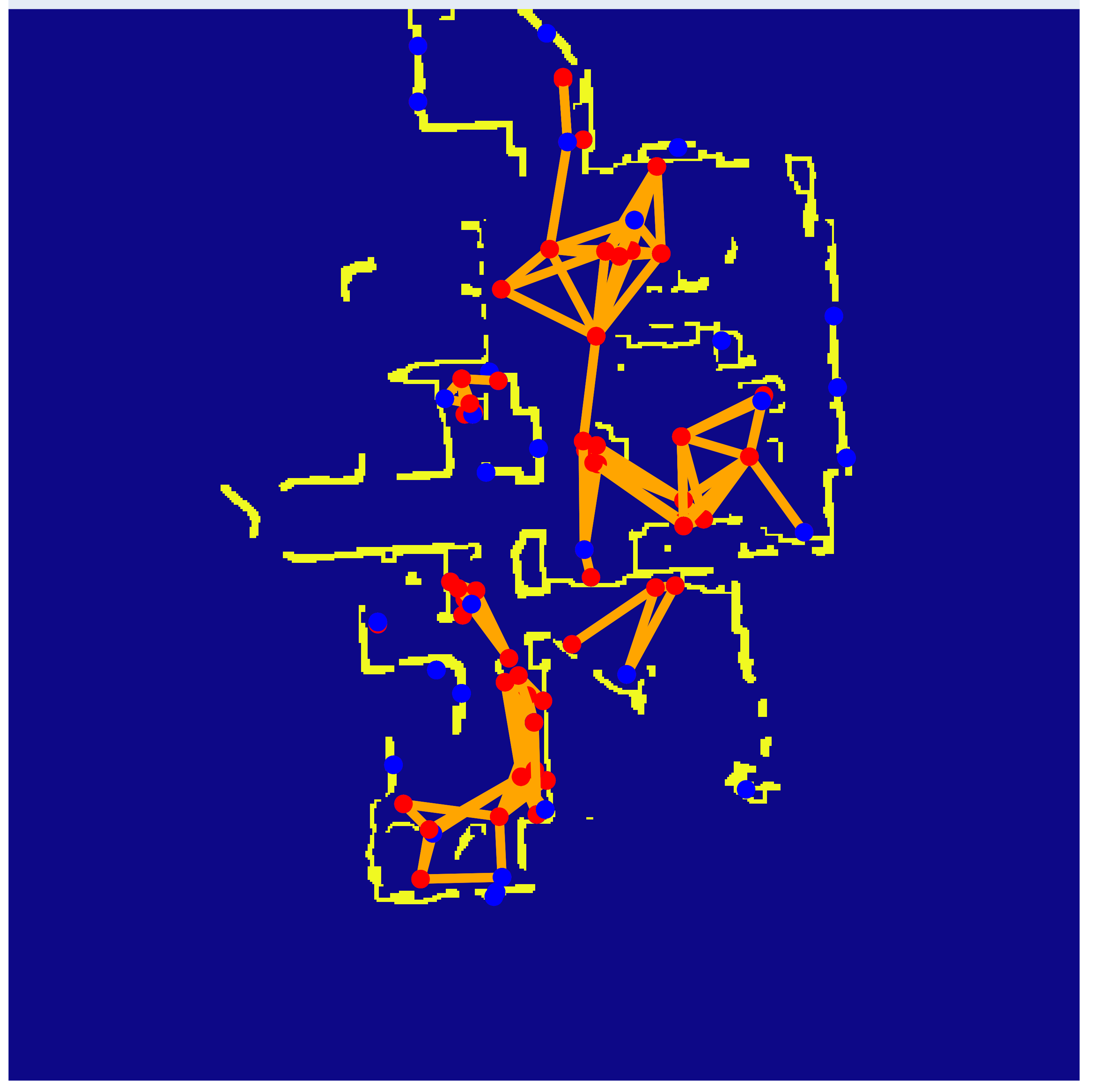}
\caption{Visualization of region groups with four sample scenes. Objects are marked by red circles, objects within the same region are connected by orange lines, and walls are indicated by yellow lines. We see that only objects that are spatially close and not separated by walls are grouped together.} 
\vspace{-3mm}
\label{fig:region_grouping}
\end{figure*}

\begin{figure}[ht!]
    \centering
    \includegraphics[width=1.0\linewidth]{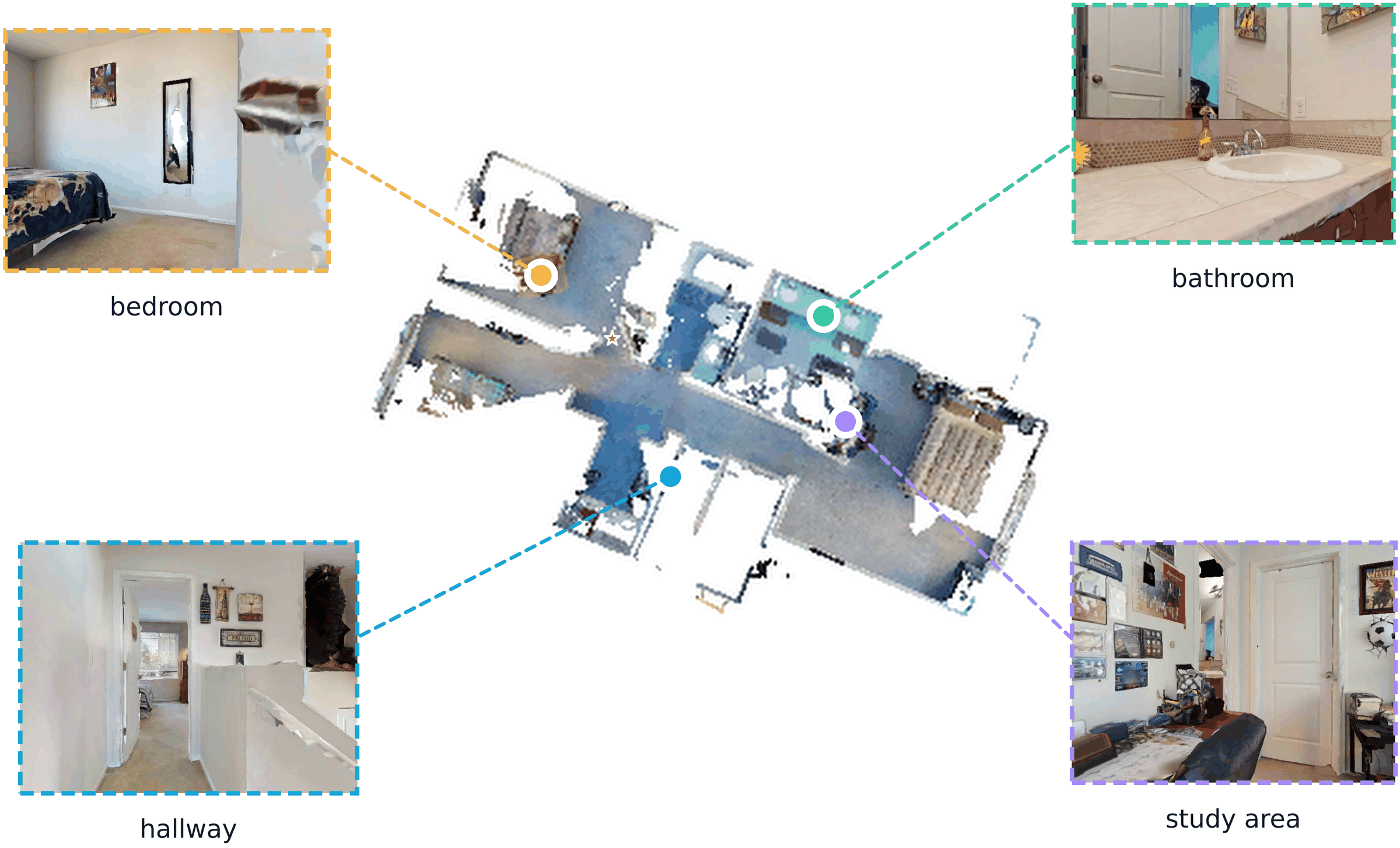}
    \vspace{-2mm}
    \caption{Visualization of region captions.}
    \vspace{-2mm}
\label{fig:region_caption}
\end{figure}

\begin{itemize}
    \item \textbf{Object detection and association.} This process updates object nodes at each time step using new visual observations. Specifically, it consists of three steps: i) detecting and segmenting 2D objects from visual observations; ii) lifting these 2D segments into 3D space using depth data to generate 3D object point clouds; and iii) associating the new 3D point clouds with existing object nodes if a match is found, or initializing them as new nodes otherwise. This ensures a comprehensive, non-redundant set of object nodes.
    \item \textbf{Region grouping.} Region grouping utilizes the geometric information of detected objects and the indoor layout to form meaningful regions, see Fig.~\ref{fig:region_grouping}. The core idea is to cluster objects that are spatially close and not separated by walls. The process involves three steps.
First, for each object $v_i$, we query a k-d tree to find its $k$ nearest neighbors. Next, we prune this neighbor set by applying two conditions: (1) the Euclidean distance satisfies $\|v_i - v_j\| < d_{\max}$, and (2) the straight-line path $traj_{i,j}$ crosses fewer than $w_{\max}$ wall pixels (i.e., $W[traj_{i,j}] < w_{\max}$). Wall pixels are identified by thresholding the gradient map at 1.2 m. To avoid connecting objects through doorways, $traj_{i,j}$ is evaluated within a narrow rectangular corridor, effectively filtering out such links. Finally, clusters are designated as regions only if they contain at least $n_{\min}$ objects, ensuring sufficient semantic information. The parameters are set as $k=5$, $d_{\max}=2.5\,\text{m}$, $w_{\max}=15$, and $n_{\min}=3$, balancing runtime efficiency and grouping accuracy.

\item \textbf{Region captioning.} For each formed region group, we extract high-level semantic context by captioning the image observations that maximally cover the grouped objects, see region caption in Fig.~\ref{fig:region_caption}. This process involves two steps. First, for each object $v_i$, we record the indices of all images in which it appears. After grouping objects into regions, we score each candidate image based on the detection confidence of the region’s constituent objects and select the top three images for each region. Second, we compute the CLIP similarity between these representative images and the set of candidate captions, assigning topK caption to each region. Note that the region candidate list can be expanded to accommodate different scenarios. Since HM3D and HSSD primarily focus on indoor home environments, we use these typical region types, including \textbf{[`living room', `study room', `dining room', `stair hall', `hallway', `bedroom', `wardrobe area', `balcony', `laundry room', `tv room', `gym', `entryway', `storage', `kitchen', `bathroom', `garage']}. Additionally, while LLMs can also generate region captions, our qualitative evaluations show that CLIP performs comparably. This is because CLIP is specifically trained for visual-text alignment, making it naturally well-suited for this captioning task. To balance effectiveness and computational cost, we therefore choose to use CLIP for generating captions.
\end{itemize}

\begin{figure*}[!ht]
\centering
\begin{subfigure}{0.3\linewidth}
    \includegraphics[width=0.99\linewidth]{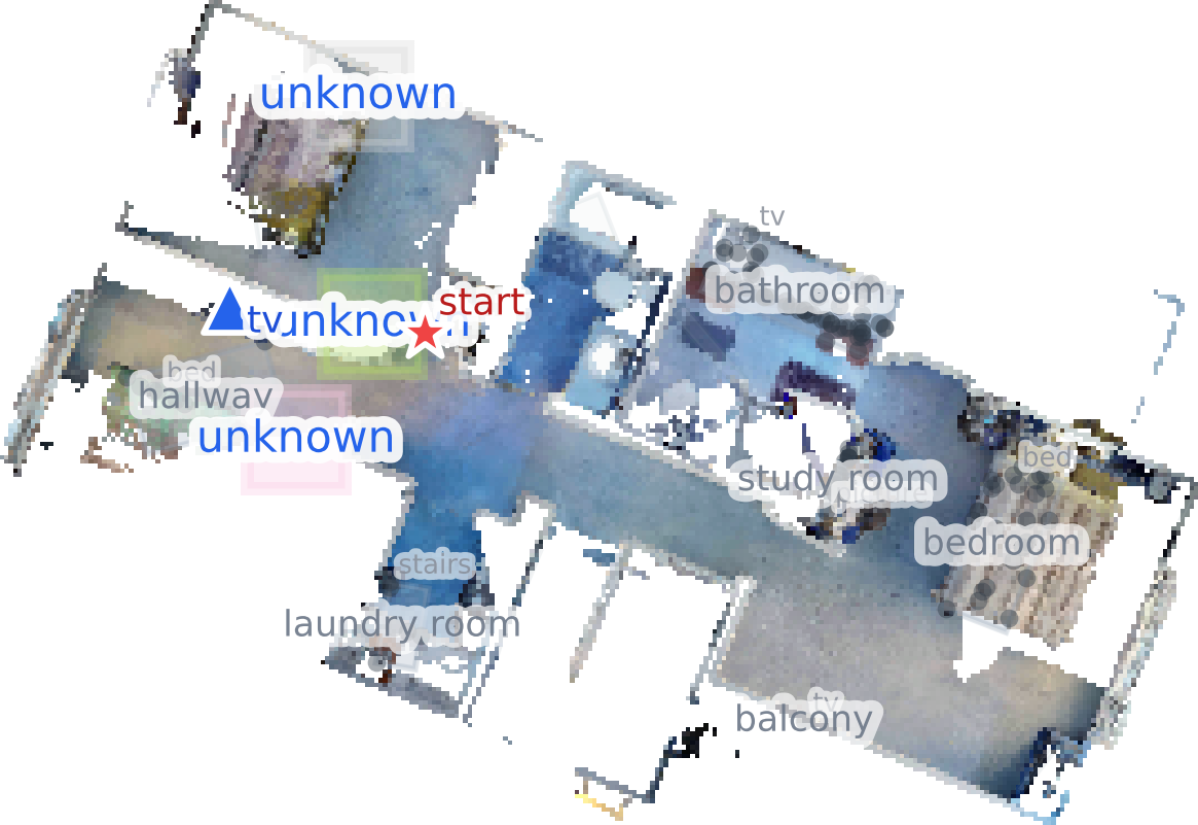}
    \vspace{-5mm}
  \caption{Observed scene graph}
  \label{Fig:bev_obs}
  \end{subfigure}
\begin{subfigure}{0.3\linewidth}
    \includegraphics[width=0.99\linewidth]{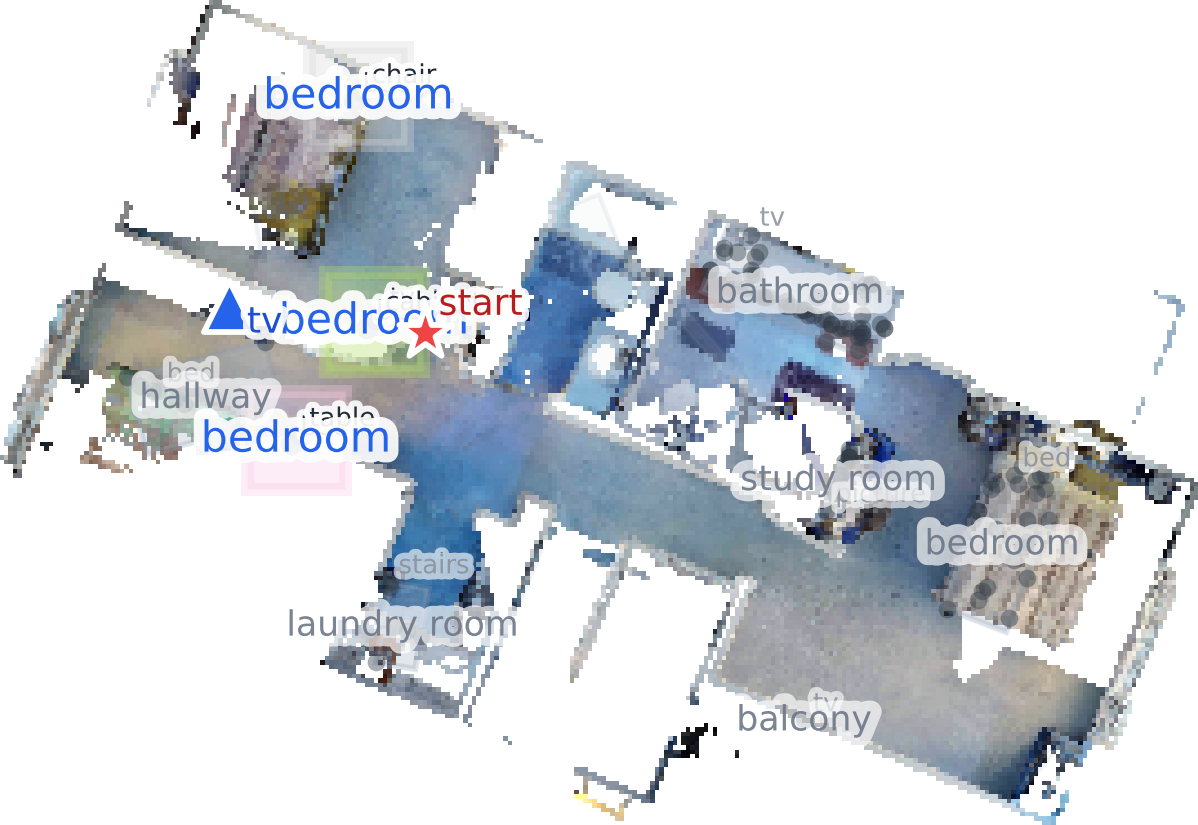}
    \vspace{-5mm}
  \caption{Predicted scene graph}
  \label{Fig:bev_pred}
  \end{subfigure}
\begin{subfigure}{0.3\linewidth}
    \includegraphics[width=0.99\linewidth]{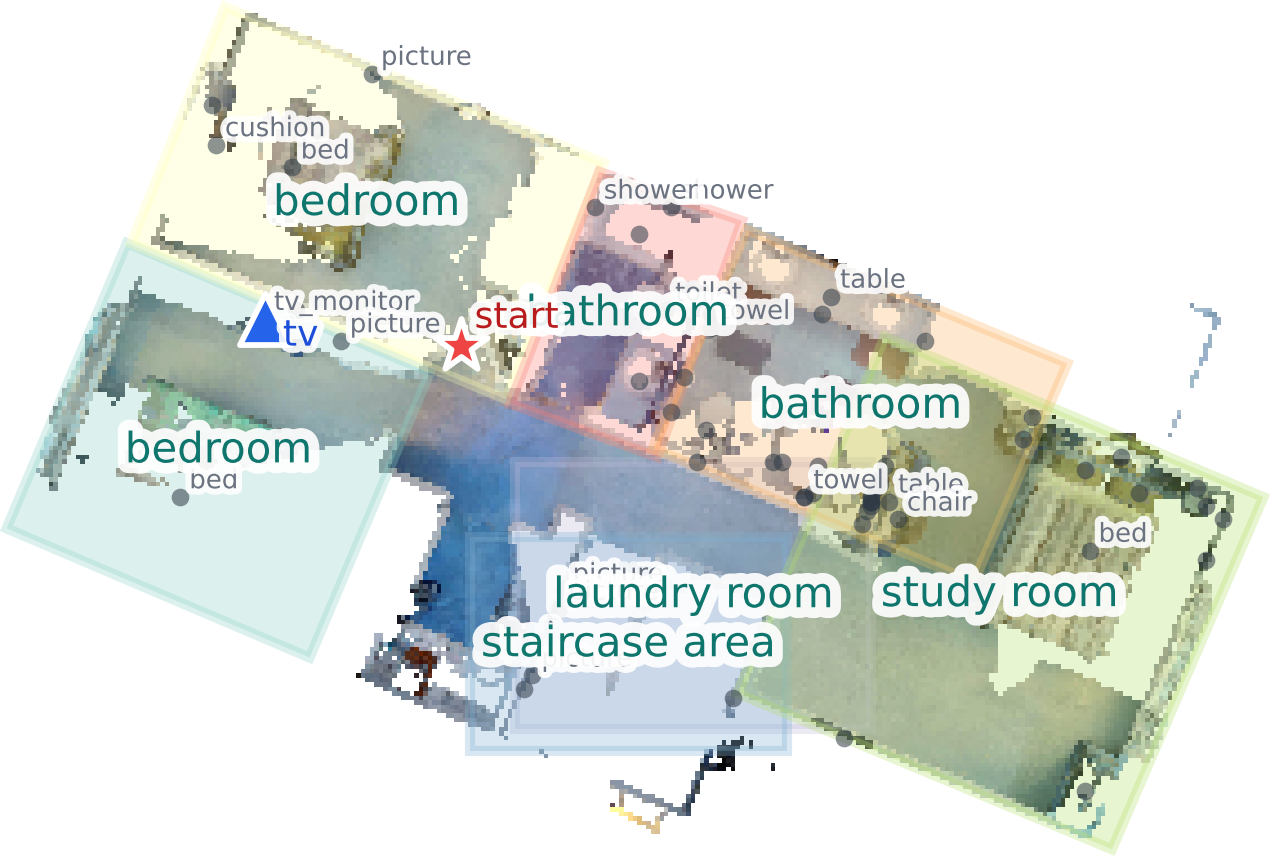}
    \vspace{-5mm}
  \caption{Ground truth scene graph}
  \label{Fig:bev_gt}
  \end{subfigure}
\caption{Visualization of scene graphs projected onto BEV image space, where objects are shown in green, observed regions in black, and predicted regions in blue. Comparing (b) and (c), the predicted scene graph accurately matches the ground-truth labels for the unknown regions, including `bathroom', `bedroom', and `living room'.} 
\vspace{-3mm}
\label{fig:bev_image}
\end{figure*}

\textbf{3. Scene graph prediction and prompts.} Given the observed scene graph, we leverage the semantic priors in VLM to predict the unexplored regions. Since VLMs inherently lack spatial reasoning, we enhance their spatial understanding by aligning textual queries and visual inputs with the target regions to provide the VLM more context at the observed regions. Building on this idea, our scene graph prediction process consists of four steps: i) identify nearby unknown regions, those adjacent to the boundaries of known areas; ii) project the observed scene graph into BEV space, generating a BEV image (see Fig.~\ref{Fig:bev_obs}); iii) construct contextual information by including nearby known regions and semantic priors such as region co-occurrence patterns; and iv) input the BEV image from step (ii) and the contextual query from step (iii) into the VLM to predict the contents of the unknown regions. The prediction prompt example is provided in Tab.~\ref{tab:region_prediction_prompts}. Fig.~\ref{fig:bev_image} compares
the observed scene graph with the predicted one. The VLM produces a more comprehensive scene graph that accurately captures the environment and provides richer semantic cues for guiding the target search. 

\begin{responselong}[%
   title={Scene graph prediction prompts},%
   label={tab:region_prediction_prompts},%
]
You are given the bird eye view of the house, and the goal is to predict what the robot might see when it explores the unknown regions (dark area) that can help find {bed}.
For each region, infer top 2 most likely captions with confidence scores (sum of two confidence scores should be 1) and also the top 3 most typical objects within the region.
HINT: Think about the typical layout of a house, here are some examples layout with possibilities:
- kitchen is usually near bathroom, living room, laundry room, study room.
- study room is usually near living room, bathroom, kitchen, hallway.
- dining room is usually near study room, storage, bathroom.
- living room is usually near bathroom, study room, laundry room, kitchen, bedroom.
- bathroom is usually near bedroom, hallway, laundry room, entryway.
- bedroom is usually near bathroom, wardrobe area, hallway, entryway.
- bedroom is usually NOT near study room, kitchen.
- dining room is usually NOT near bedroom.
- study room is usually NOT near bedroom.
- living room is usually NOT near storage, wardrobe area.
**Available region choices**: living room, study room, dining room, stair hall, hallway, bedroom, wardrobe area, balcony, laundry room, tv room, gym, entryway, storage, kitchen, bathroom, garage, unknown. Set caption to 'unknown' if you are uncertain or no choice makes sense.
**Unknown regions locations and nearby regions**:
Unknown region 0, center: [345 332], nearby regions:
None 
----------
Unknown region 1, center: [251 343], nearby regions:
Region 6: {'bedroom': 0.232177734375, 'laundry room': 0.2099609375} center: [239.29270935058594, 306.1341552734375] contained objects: ['cabinet', 'cabinet', 'cabinet', 'cabinet', 'cabinet', 'tv', 'cabinet', 'cabinet', 'cabinet'] 
Region 20: {'bedroom': 0.50146484375, 'hallway': 0.49169921875} center: [211.0, 377.0] contained objects: ['picture'] 
Region 5: {'wardrobe area': 0.45751953125, 'bedroom': 0.449462890625} center: [216.0, 299.0] contained objects: ['cabinet', 'cabinet', 'cabinet', 'cabinet'] 
----------
Unknown region 2, center: [308 382], nearby regions:
None 
----------
Output Requirements:
- DO NOT include the observed regions above, ONLY predict the unknown regions.
- DO NOT explain your task, just give the short reasoning and the final prediction.
- The scene graph must follow the given JSON structure. REMOVE spaces in the JSON string.
- Including a start flag of "Start" and an end flag "End", in between is the final scene graph.
- Give a short 20-words-max reasoning for each predicted region.
Output Format:
# Start
```json
```
# End
Example: 
dict('regions': [dict('id': pred_#RegionID, 'caption': dict(#RegionType1: #ConfScore1, #RegionType2: #ConfScore2), 'reasoning': #Reasoning, 'center': [x, y], 'objects': [dict('caption': #ObjectType, 'center': [x, y], 'confidence': #ConfScore, 'corr_score': #CorrelationWithTarget)])])])
\end{responselong}

\textbf{4. Update mechanism.} The scene graph updates dynamically alongside new observations. To balance accuracy and efficiency, its components update at different frequencies:

\begin{itemize}
    \item Scene graph generation: Updates synchronously with each visual observation.
    \item Scene graph prediction: Updates asynchronously (at a lower frequency) to save computation. It is triggered either every 20 timesteps or when a major environmental change occurs (e.g., a region node changes or a new region is discovered).
\end{itemize}

The hierarchy of the scene graph updates from the bottom up:

\begin{itemize}
    \item \textbf{Object level}. New visual inputs trigger the object detector, and object association generates or updates the corresponding object nodes.
    \item \textbf{Region level}. The appearance of new object nodes triggers region grouping and captioning to update the region nodes.
    \item \textbf{Floor level}. Floor nodes are updated when a new stair object node is identified.
    \item \textit{Prediction level}. Triggered based on the lower-frequency prediction conditions.
\end{itemize}

\textbf{5. Handling of hallucinated predictions.}
We mitigate the impact of hallucination at three levels.
\begin{itemize}
    \item \textbf{Prediction design.} We constrain the prediction context to the unknown region surrounding each frontier, rather than predicting the entire unexplored space, and we predict only coarse region categories and key objects instead of complete, fine-grained scene graphs. This narrower scope substantially reduces the room for hallucination.
    \item \textbf{Prediction usage.} The predicted scene graph is used only for target gain estimation and is never merged into the observed scene graph, so prediction errors cannot propagate to subsequent steps. Once a region is observed, its prediction is discarded and replaced by the observed scene graph. Imagined regions never overwrite the observed graph or the navigation map; they serve solely as an auxiliary semantic prior for frontier selection.
    \item \textbf{Navigation system.} Our exploitation-exploration switching mechanism further filters noisy predictions: imagined regions influence navigation only when they are highly relevant to the target; otherwise, the agent falls back to the observed scene graph and geometric exploration cues.
\end{itemize}

\subsection{3D planner}

The 3D planner receives the designated location and generates the next discrete action. The agent executes this action in the environment and receives new observations, completing the full update loop for the timestep. 

\textbf{1. The 3D planner takes as input either the verified goal position or the next frontier.} These two input options correspond to the two pipelines the system can route to.

\begin{itemize}
    \item \textbf{Goal found}: If the target goal matches an existing object node, the Context-Aware Goal Verification module passes the object's 3D location directly to the 3D planner.
    \item \textbf{Goal search}: If the goal is not yet found, the scene graph is passed to the Target Information Gain Estimation module to select the next best frontier. This frontier location is then sent to the 3D planner. 
\end{itemize}

Specifically, Fig. 4 illustrates that selecting the next frontier depends on both the semantic (exploitation) gain from the LLM reasoning and the geometric (exploration) gain from the raycasting-based occupancy map. Specifically, we compute the accumulated exploration gain in three steps. First, for each frontier, we generate a sequence of waypoints from the agent's current position using a Fast Marching Method-based planner. To improve computational efficiency, the resulting path is uniformly subsampled to at most $n<12$ waypoints, always including the start (current position) and end (frontier) points. Second, for each waypoint $\{(x_i, y_i)\}_{i=1}^n$, we use a raycasting algorithm to cast 20 rays of length $r_{\text{ray}}$ to detect the first intersections with occupied cells in the occupancy map. Note that the ray number is set as 20 to balance the grid coverage and the efficiency. This yields a set of visible vertices for each waypoint, denoted as $(x'_{i1}, y'_{i1}), (x'_{i2}, y'_{i2}), \dots, (x'_{i20}, y'_{i20})$, forming a polygon that approximates the visible area from that waypoint. Each polygon is rasterized into a grid set $P_i$. Third, exploration gain is computed iteratively using the unknown map $U$. Let $A_1 = P_1 \cap U$, $A_2 = (P_2 \cap U) \setminus A_1$, $A_3 = (P_3 \cap U) \setminus (A_1 \cup A_2)$, and so on, such that $A_i$ represents disjoint sets of unknown cells first observed from the $i$-th waypoint. The final exploration gain is computed as the sum of disjoint newly observed regions, given by
$\frac{1}{n\pi r_{\text{ray}}^2} \sum_{i=1}^n \gamma^{i-1} |A_i|,$
where $\gamma$ is a discount factor. The normalization term $\frac{1}{n\pi r_{\text{ray}}^2}$ ensures the gain is scaled between 0 and 1, making it a bounded ratio. Note that the discount factor $\gamma$ is used to weight the visible regions based on their proximity regions closer to the agent are more likely to be reached and thus contribute more to the overall reward.

\begin{figure*}[!ht]
\centering
\begin{subfigure}{0.3\linewidth}
    \centering
    \includegraphics[width=0.99\linewidth]{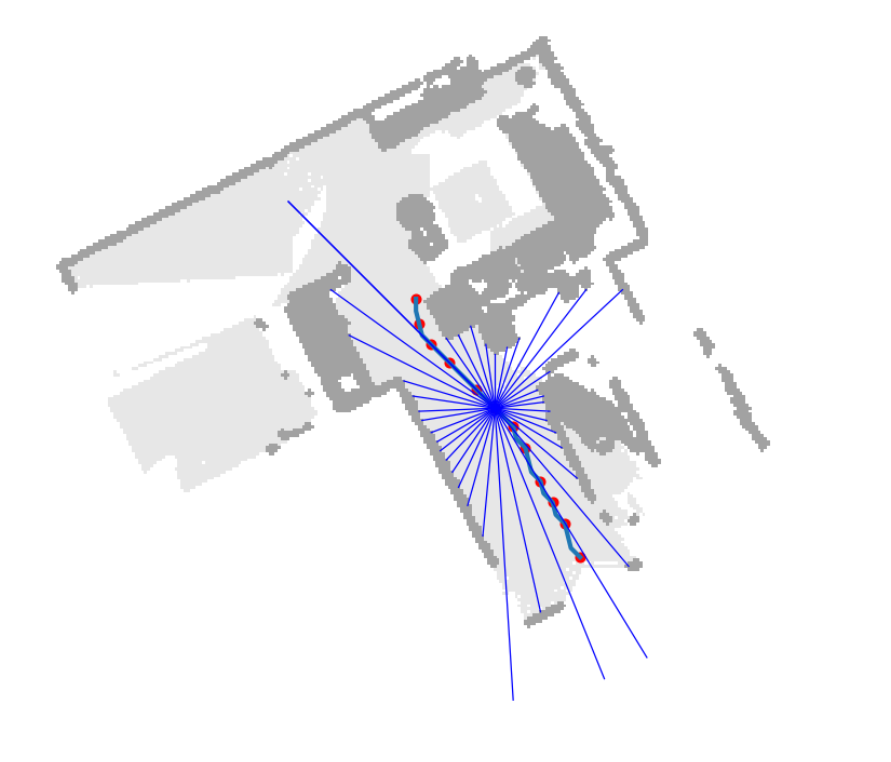}
    \vspace{-5mm}
  \caption{Raycasting at each waypoint}
  \label{Fig:single_point_raycast}
  \end{subfigure}
\begin{subfigure}{0.3\linewidth}
    \centering
    \includegraphics[width=0.99\linewidth]{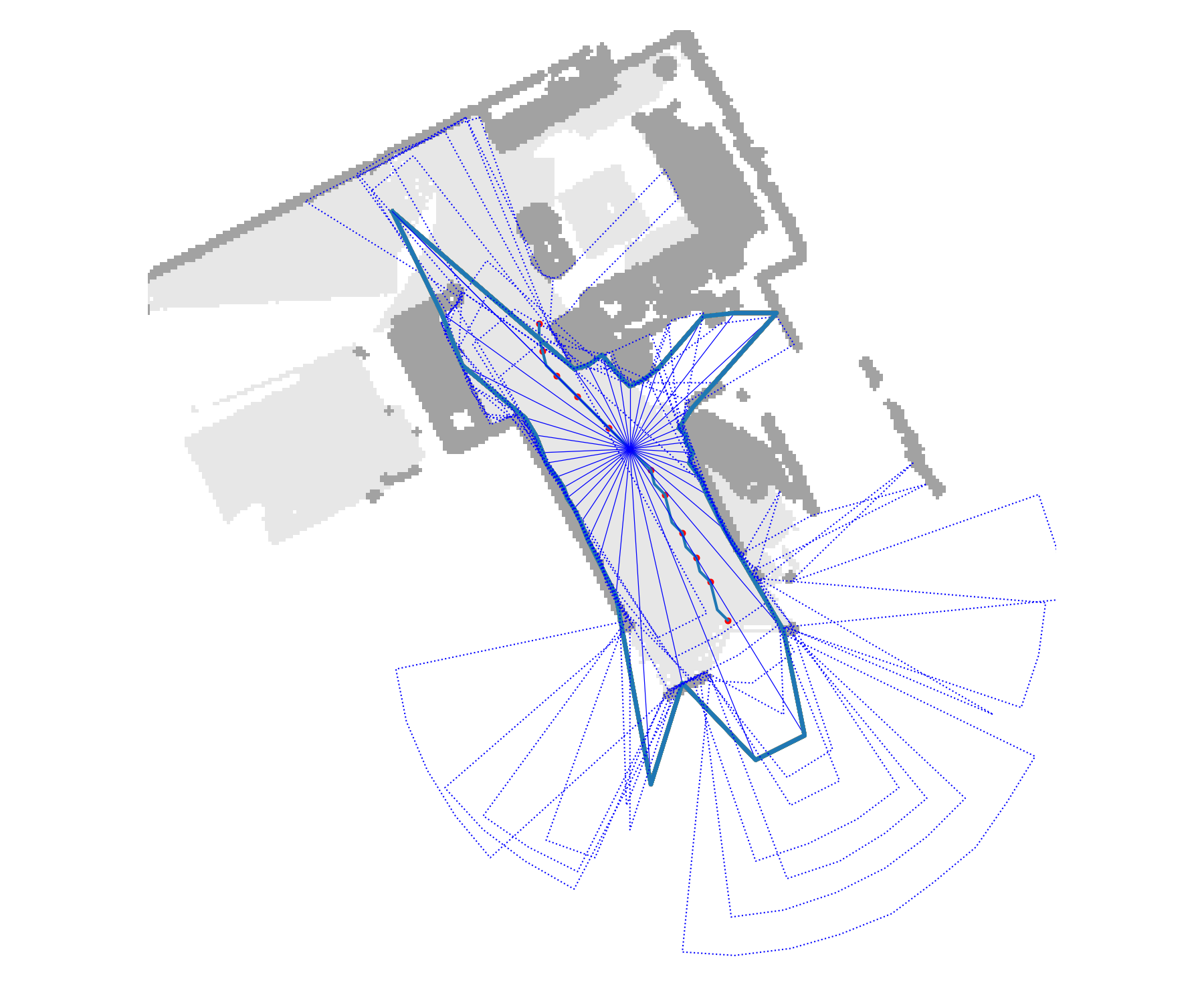}
    \vspace{-5mm}
  \caption{Approximate via polygons}
  \label{Fig:visible_raster}
  \end{subfigure}
\begin{subfigure}{0.3\linewidth}
    \centering
    \includegraphics[width=0.99\linewidth]{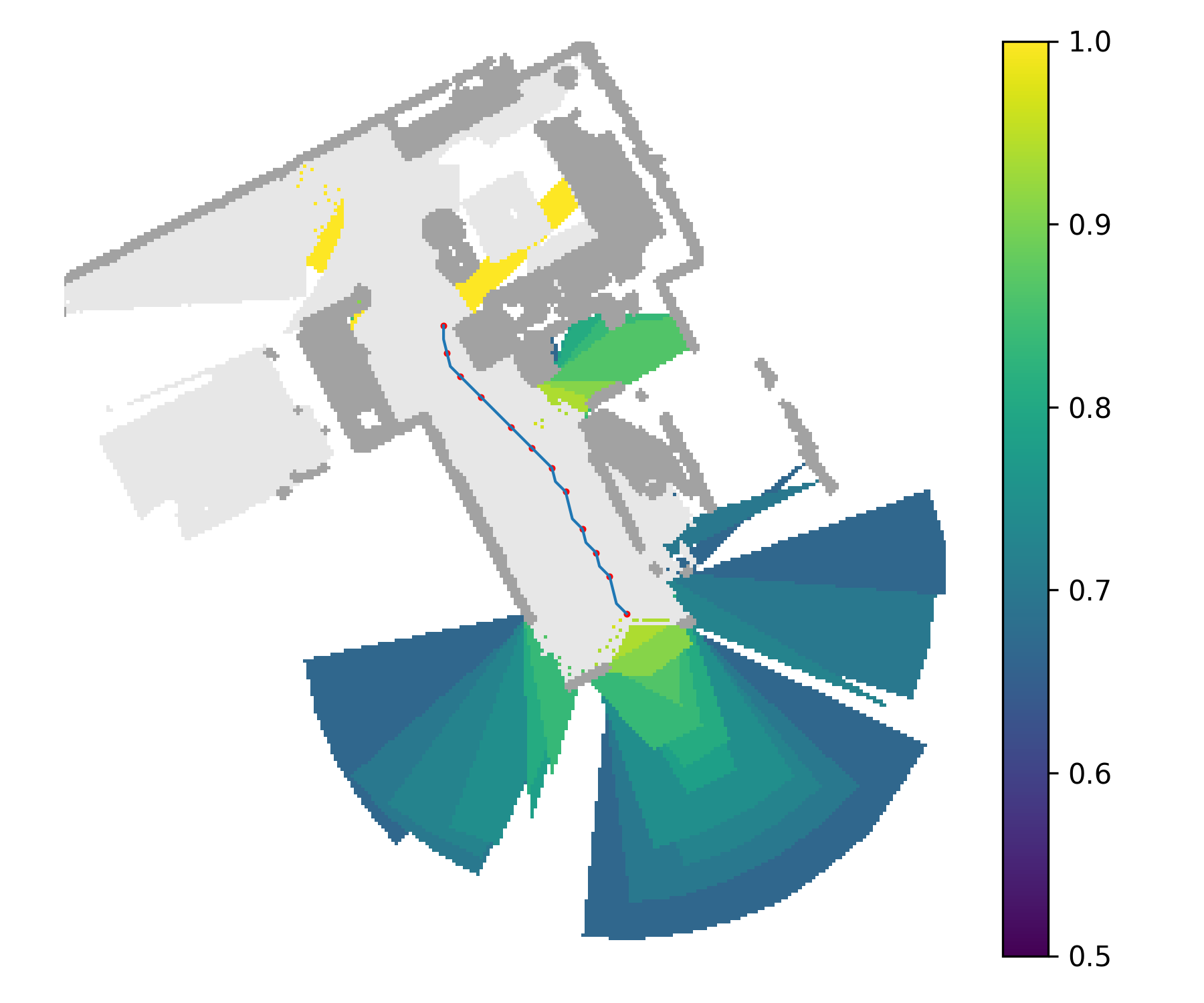}
    \vspace{-5mm}
  \caption{Discounted visible unknown}
  \label{Fig:path_raycast}
  \end{subfigure}
\caption{The exploration gain is calculated as the accumulated visible unknown space that the agent may explore on its path towards each frontier.} 
\vspace{-3mm}
\label{Fig:raycast}
\end{figure*}

\textbf{2. The 3D planner employs the Fast Marching Method (FMM)~\cite{sethian1999fast}, following~\cite{yokoyama2024vlfm}, to generate the next action.} Unlike conventional FMM, which is limited to flat 2D surfaces, our approach reuses FMM for both cross-floor and inter-floor navigation. 

To extend FMM from traditional 2D single-floor navigation to 3D path planning and traverse between floors, the key idea is to detect stair regions and represent them as traversable areas, allowing the planner to generate paths over stairs. Specifically, our approach involves three steps. First, we identify candidate stair objects using object detection. Next, we apply a 2.5D BEV mapping strategy to mark these stair regions as navigable within the map. Finally, a stair climbing policy is triggered to guide the agent seamlessly across different floors.

\begin{figure*}[!t]
\centering
\begin{subfigure}{0.24\linewidth}
    \centering
    \includegraphics[width=0.99\linewidth]{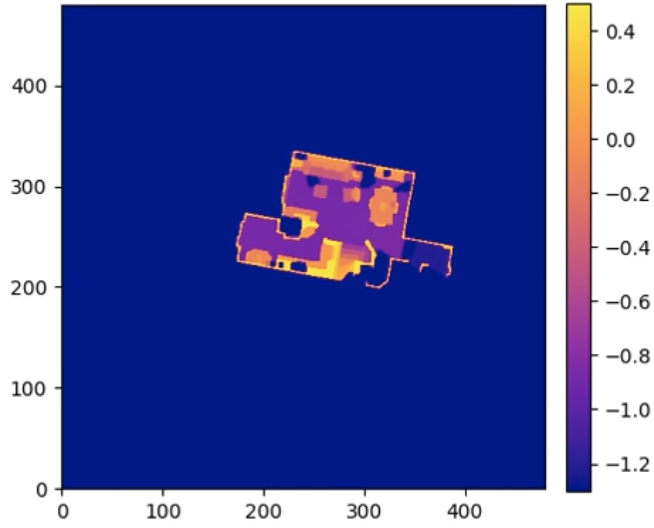}
    \vspace{-5mm}
  \caption{Height map}
  \label{Fig:height_map}
  \end{subfigure}
\begin{subfigure}{0.24\linewidth}
    \centering
    \includegraphics[width=0.99\linewidth]{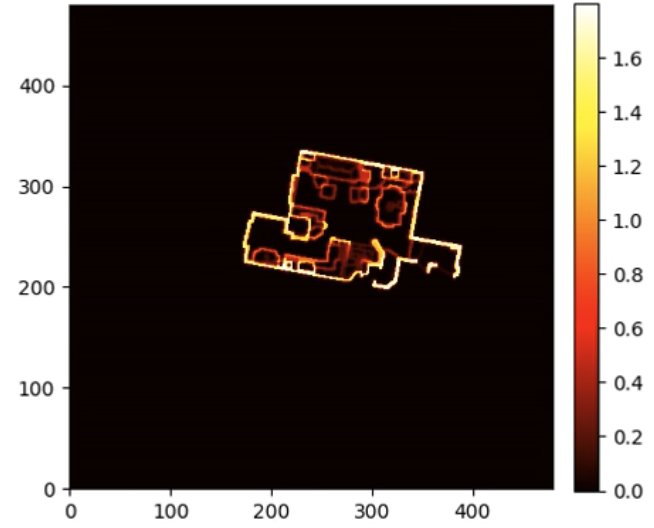}
    \vspace{-5mm}
  \caption{Gradient map}
  \label{Fig:gradient_map}
  \end{subfigure}
\begin{subfigure}{0.24\linewidth}
    \centering
    \includegraphics[width=0.99\linewidth]{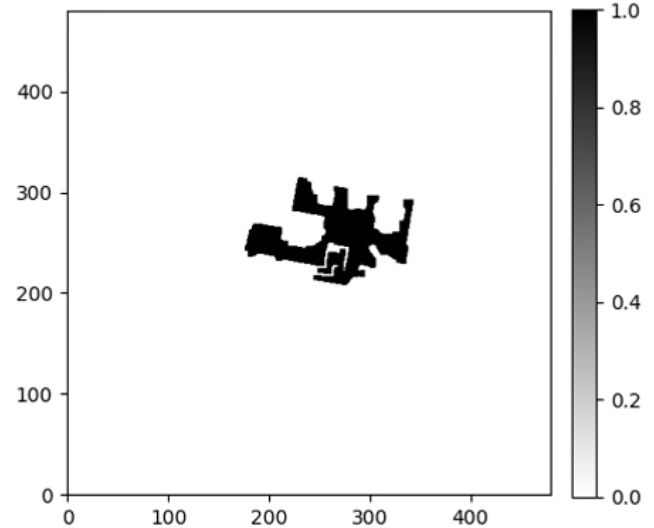}
    \vspace{-5mm}
  \caption{Traversable map}
  \label{Fig:traversible_map}
  \end{subfigure}
\begin{subfigure}{0.24\linewidth}
    \centering
    \includegraphics[width=0.99\linewidth]{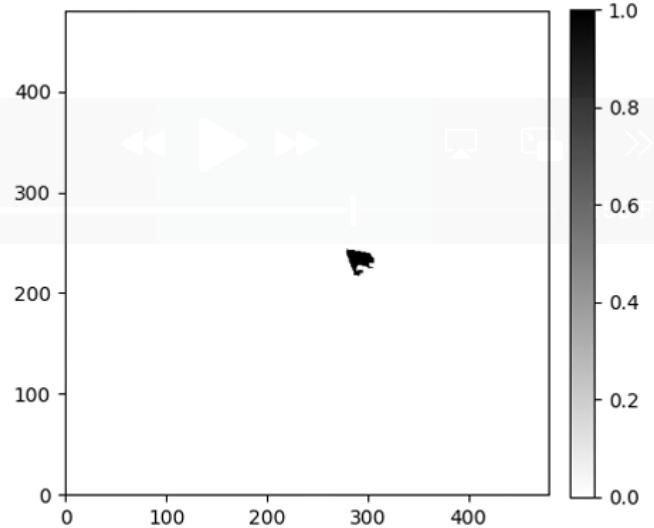}
    \vspace{-5mm}
  \caption{Stair map}
  \label{Fig:stair_map}
  \end{subfigure}
\caption{Visualization of internal maps used to construct a specialized traversable map that accounts for stairs.} 
\vspace{-4mm}
\label{Fig:multifloor_maps}
\end{figure*}

\begin{itemize}
    \item \textbf{Stair detection.} To support multi-floor navigation, we append \textit{stairs} as object of interest to the Grounded-SAM detector. Once a stair instance is detected, we classify its direction, upward or downward, based on the it relative height derived from its point cloud. If the average height is greater than the agent’s base height ($0.1$m), the stair is added to the \textit{upstair\_candidate\_list}; otherwise, it is added to the \textit{downstair\_candidate\_list}. When no frontiers remain on the current level, the stair-climbing module is triggered. At this point, the agent constructs a $2.5$D map and attempts to transition to another floor if valid stair candidates are found. We detail the $2.5$D map construction below.
    \item \textbf{Stair traversable map generation.} Traditional 2D occupancy maps treat stairs as obstacles due to the absence of height information. To overcome this limitation, we enhance the map with height-awareness to create a $2.5$D representation. When the stair-climbing state is triggered, all point clouds from the current floor are projected into a 2D height map (Fig.\ref{Fig:height_map}). We then compute a gradient map using a customized kernel to estimate height changes in meters (Fig.\ref{Fig:gradient_map}). The gradient map highlights environmental edges; pixels with near-zero gradients correspond to flat floors, while those below a certain threshold are classified as stairs, resulting in a traversable map (Fig.\ref{Fig:traversible_map}). Based on the Habitat simulator’s maximum climbable height of $0.2$m, we set the threshold heuristically to $0.3$ to accommodate discretization noise. Detected stair regions are further emphasized in a stair map (Fig.\ref{Fig:stair_map}), allowing the agent to identify and traverse them.
    \item \textbf{Stair climbing.} To transition between floors, the agent follows a 3-stage procedure: i) approach the stair entrance, ii) ascend or descend the stairs, and iii) confirm arrival on a new floor. In the first stage, if the agent decides to climb upstairs, the lowest point of an upstairs stair candidate is identified as the \textit{stair\_entrance} and used as the goal. The FMM planner then generates a path to reach it. Upon arrival, the agent enters the stair-climbing state. In this stage, it selects the highest reachable point within a $2\times2$ meters local map window as the next goal. Since the stairs are now marked as traversable in the $2.5$D map, FMM planning proceeds smoothly. In the Habitat simulator, if the elevation difference is within the allowed climbing limit, a forward motion action automatically elevates the agent. While climbing, the agent continuously monitors the number of points within $\pm0.1$m of its base height. These points are treated as a floor plane. If the number of such floor points exceeds 800, the agent considers itself to have reached a new level and transitions to the third stage. It then resumes exploration on the new floor.
\end{itemize}

\subsection{Comparison with VLA}

Here, we clarify the unique technical novelty of SGImagineNav relative to VLA systems from both conceptual and practical perspectives.

\textbf{1. Conceptual comparison and novelty}
Conceptually, the core distinction and novelties of \texttt{SGImagineNav} relative to standard end-to-end VLA navigation systems lie in three key dimensions:

\begin{itemize}
    \item \textbf{Proactive exploration via generative world imagination.} Standard VLA models typically generate low-level control actions reactively based on current observations and histories. In contrast, \texttt{SGImagineNav} introduces a proactive mechanism that leverages semantic priors within a VLM to predict and ``imagine'' unseen environment topologies (i.e., completing the semantic scene graph). This proactive  capability is orthogonal to standard VLA architectures and could theoretically be integrated into VLA pipelines to enhance their long-term planning.
    \item \textbf{Long-horizon scaling via symbolic world models.} VLA-style systems are heavily bottlenecked by the context window limitations of Transformers, making it computationally prohibitive to maintain high-fidelity visual memories over long spatiotemporal footprints. \texttt{SGImagineNav} mitigates this by abstracting the environment into a structured, symbolic hierarchical scene graph. This compact representation provides a global context, enabling efficient long-horizon navigation in large-scale real-world environments.
    \item \textbf{Explainability and controllability.} End-to-end VLAs function largely as black boxes, making them difficult to diagnose or control. Our framework decouples navigation into distinct semantic and geometric information gains. This explicit separation provides high interpretability, enabling finer control over the agent's behavior. For instance, it allows the integration of a meta-policy to adaptively balance exploitation and exploration weights conditioned on the environment's familiarity.
\end{itemize}

\textbf{2. Empirical Comparison and Evaluation.}
VLA models often struggle with long-horizon navigation due to high memory overhead, hallucinations, and a lack of geometric mapping. To verify this, we evaluate our method against representative VLA baselines (e.g., UniNavid) in large-scale scenes within Isaac Sim, presenting both quantitative and qualitative comparisons. 

\begin{figure}[ht!]
    \centering
    \includegraphics[width=1.0\linewidth]{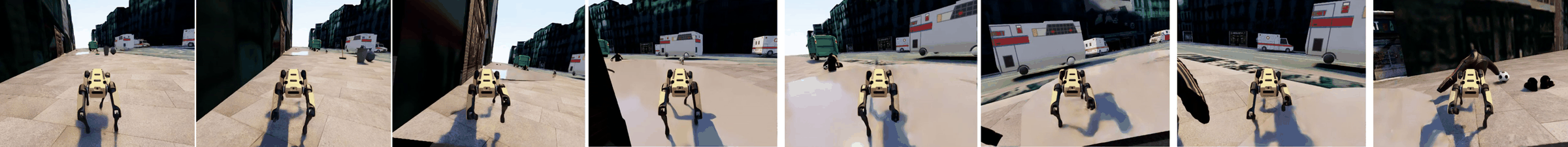}
    \vspace{-2mm}
    \caption{Example of SGImagineNav in Isaac Sim large-scale scenario.}
    \vspace{-2mm}
\label{fig:navverse_example_sg_vln}
\end{figure}

\begin{figure}[ht!]
    \centering
    \includegraphics[width=1.0\linewidth]{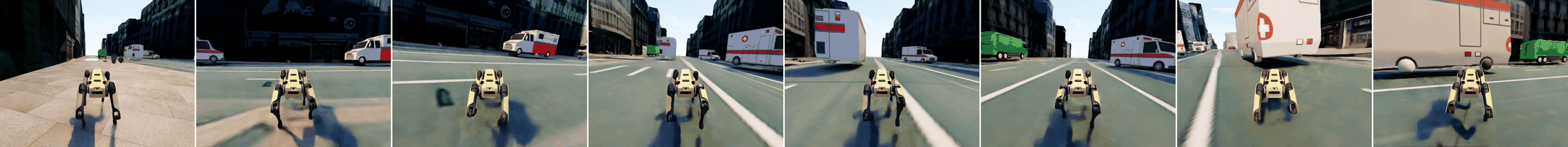}
    \vspace{-2mm}
    \caption{Example of UniNavid in Isaac Sim large-scale scenario.}
    \vspace{-2mm}
\label{fig:navverse_example_uninavid}
\end{figure}

\begin{figure}[ht!]
    \centering
    \includegraphics[width=0.5\linewidth]{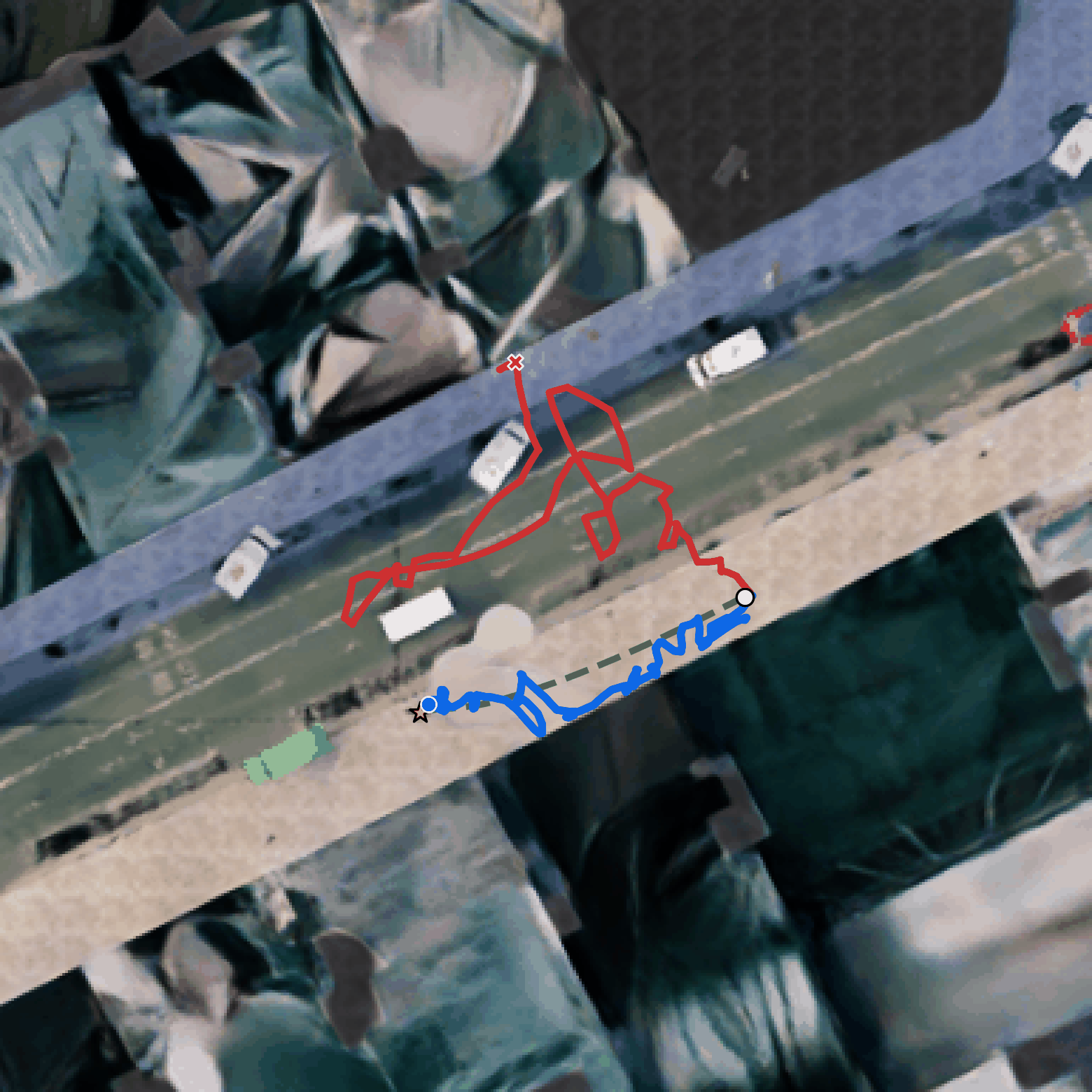}
    \vspace{-2mm}
    \caption{Trajectories of SGImagineNav and UniNavid example in Isaac Sim large-scale scenario.}
    \vspace{-2mm}
\label{fig:navverse_example_bev}
\end{figure}

\begin{itemize}
    \item \textbf{Quantitatively}, Tab.~\ref{tab:navigation_results} demonstrates that \texttt{SGImagineNav} yields substantially higher SR and SPL in large-scale outdoor environments, where target localization demands structured exploration and explicit memory. In contrast, VLA baselines perform well primarily in short-range tasks with obvious visual cues, such as in HM3D household scenes. 
    \item \textbf{Qualitatively}, the trajectory comparison in Fig.~\ref{fig:navverse_example_bev} highlights that lacking geometric mapping or consistent memory causes the VLA baseline to suffer from cyclic looping. Conversely, \texttt{SGImagineNav} successfully uncovers new regions by tracking semantic clues (see Fig.~\ref{fig:navverse_example_sg_vln} and Fig.~\ref{fig:navverse_example_uninavid} for detailed observations).
\end{itemize}

\begin{table}[!ht]
\centering
\caption{SGImagineNav outperforms VLA-style system in large-scale outdoor scenarios.}
\label{tab:navigation_results}
\begin{tabular}{lcc}
\toprule
\multirow{2}{*}{Method} & \multicolumn{2}{c}{Large-Scale Scenario} \\
\cmidrule(r){2-3}
 & SR (\%) & SPL (\%) \\
\midrule
UniNaVid (VLA) & 4.95 & 0.89 \\
SGImagineNav & 8.91 & 1.06 \\
\bottomrule
\end{tabular}
\end{table}

\subsection{Efficiency and sensitivity analysis}

To evaluate the computational efficiency, we detail the average runtime, inference frequency, and number of VLM/LLM calls per episode in Tab.~\ref{tab:computational_cost}. Additionally, a detailed runtime breakdown is provided in Tab.~\ref{tab:runtime_breakdown}. Note that both the VLM and LLM utilize API calls. All reported runtimes were measured on a workstation with an Intel Xeon Platinum 8559C CPU and an NVIDIA RTX PRO 6000 Blackwell GPU. The language model used in our pipeline is Qwen2.5-7B. From these results, we observe the following:
\begin{itemize}
    \item \textbf{Navigation decisions}: The navigation decision updates at 1.2 Hz, where the primary bottlenecks are object detection and exploration gain estimation. To improve latency, object detection can be parallelized to run at a higher frequency rather than sequentially. Furthermore, the raycasting for exploration gain estimation is currently implemented on the CPU; migrating this process to the GPU would significantly increase processing speed.
    \item \textbf{Prediction frequency}: Prediction occurs at a much lower frequency, approximately once every 20 steps (or under 5 meters). This lower frequency is justified because predictions only need to be triggered when significant visual changes occur. Given that each API call takes approximately 6 seconds to execute, this targeted prediction frequency effectively balances model effectiveness with overall system efficiency.
\end{itemize}

\begin{table}[!ht]
\centering
\caption{Computational cost and inference statistics per episode for SGImagineNav. With an average episode length of approximately 300 steps, the recorded per-step inference time accounts for the entire pipeline.}
\label{tab:computational_cost}
\begin{tabular}{lc}
\toprule
\textbf{Metric} & \textbf{Value} \\
\midrule
Avg. Runtime (s)       & 605.9 $\pm$ 378.8 \\
Basic Freq. (Hz)       & 1.20              \\
Prediction Freq. (Hz)  & 0.02              \\
VLM Calls              & 17.0 $\pm$ 10.6   \\
LLM Calls              & 18.2 $\pm$ 1.1    \\
Memory (GB)            & 11.8              \\
\bottomrule
\end{tabular}
\end{table}

\begin{table}[!ht]
\centering
\caption{Runtime breakdown of the agent decision step: object detection, mapping, scene graph update, exploration and exploitation information gain estimation, and planning.}
\label{tab:runtime_breakdown}
\begin{tabular}{lcc}
\toprule
Module & Time (s) & Percentage (\%) \\
\midrule
Object detection & 0.332 & 39.8 \\
Mapping & 0.095 & 11.3 \\
Scene graph update & 0.089 & 10.6 \\
Exploitation gain & 0.024 & 2.9 \\
Exploration gain & 0.225 & 27.0 \\
3D planner & 0.070 & 8.4 \\
\midrule
Total & 0.834 & 100.0 \\
\bottomrule
\end{tabular}
\end{table}

To evaluate the robustness of our system, we conduct a sensitivity analysis by varying the threshold over a wide range. The results, evaluated on the same 400 random episodes used for our ablation studies, are presented in Tab.~\ref{tab:threshold_ablation} and included in the Appendix of the revised manuscript. Note that, the system selects the exploitation frontier only when the maximum semantic exploitation gain exceeds the threshold, signaling a high probability of locating the target; otherwise, it prioritizes the exploration frontier to uncover large unknown regions. Consequently, \textbf{a lower threshold implies greater trust in semantic cues (favoring exploitation), whereas a higher threshold biases the agent toward exploration.}

\begin{table}[!ht]
\centering
\caption{Sensitivity analysis of the frontier selection threshold across different scenarios. }
\label{tab:threshold_ablation}
\begin{tabular}{ccc}
\toprule
Threshold ($\tau$) & SR (\%) & SPL (\%) \\ 
\midrule
0.1 & 61.25 & 25.32 \\
0.2 & \underline{63.50} & 28.75 \\
0.3 & \textbf{65.25} & \textbf{31.41} \\
0.4 & 62.50 & \underline{30.34} \\
0.5 & 62.00 & 29.92  \\
0.6 & 61.50 & 29.62 \\
0.8 & 61.25 & 29.16  \\
1.0 & 62.00 & 28.92 \\
\bottomrule
\end{tabular}
\end{table}

Our analysis yields two primary insights: 
\begin{itemize}
    \item \textbf{High robustness:} SGImagineNav is highly robust within an intermediate threshold window ($\tau \in [0.2, 0.4]$), where both SR and SPL fluctuate by less than 2\%.
    \item \textbf{Optimal trade-off:} An intermediate threshold balances exploration and exploitation, leading to peak overall performance across most scenarios. Specifically, our default threshold of $0.3$ generalizes well across structurally diverse environments without requiring manual, per-environment tuning.
\end{itemize}

\end{document}